\documentclass[10pt,journal,compsoc]{IEEEtran}
\ifCLASSOPTIONcompsoc
  \usepackage[nocompress]{cite}
\else
  \usepackage{cite}
\fi
\ifCLASSINFOpdf
\else
\fi
\usepackage{subcaption}
\usepackage{hyperref}
\usepackage{graphicx}
\usepackage{xcolor}
\usepackage{gensymb}
\usepackage[export]{adjustbox}
\usepackage{amsmath}
\usepackage[hang]{footmisc}
\usepackage{amssymb}
\usepackage{float}
\usepackage{lipsum}

\begin{document}
\title{Pose Impact Estimation on Face Recognition using 3D-Aware Synthetic Data with Application to Quality Assessment}
% lists the Biometrics Council journals use for "first footnote" author
\author{Marcel~Grimmer,
        Christian~Rathgeb,
        Christoph~Busch % <-this % stops a space
\IEEEcompsocitemizethanks{\IEEEcompsocthanksitem M. Grimmer is with the Norwegian Biometrics Laboratory (NBL), Norwegian University of Science and Technology (NTNU), Gj{\o}vik, Norway.\protect\\
E-mail: marceg@ntnu.no
\IEEEcompsocthanksitem C.Rathgeb and C. Busch are with the Biometrics and Internet-Security Research Group (da/sec), Hochschule Darmstadt (h-da), Darmstadt. Germany.\protect\\
E-mail: christian.rathgeb@h-da.de, christoph.busch@h-da.de

}% <-this % stops an unwanted space
}

\IEEEtitleabstractindextext{%
\begin{abstract}
Evaluating the quality of facial images is essential for operating face recognition systems with sufficient accuracy. The recent advances in face quality standardisation (\textit{ISO/IEC CD3 29794-5}) recommend the usage of component quality measures for breaking down face quality into its individual factors, hence providing valuable feedback for operators to re-capture low-quality images. In light of recent advances in 3D-aware generative adversarial networks, we propose a novel dataset, \textit{Syn-YawPitch}, comprising $1,000$ identities with varying yaw-pitch angle combinations. Utilizing this dataset, we demonstrate that pitch angles beyond 30 degrees have a significant impact on the biometric performance of current face recognition systems. Furthermore, we propose a lightweight and explainable pose quality predictor that adheres to the draft international standard of ISO/IEC CD3 29794-5 and benchmark it against state-of-the-art face image quality assessment algorithms. 
\end{abstract}

\begin{IEEEkeywords}
Face Recognition, Error Analysis, Synthetic Data, 3D-Aware Face Image Synthesis
\end{IEEEkeywords}}

\maketitle

\IEEEdisplaynontitleabstractindextext

% \fi
\IEEEpeerreviewmaketitle

\IEEEraisesectionheading{\section{Introduction}\label{sec:introduction}}

\IEEEPARstart{A}{cross} the globe, face recognition (FR) systems have become increasingly prevalent in our everyday life. In the European Union, sensitive applications are equipped with biometric recognition, such as the \textit{Entry-Exit System (EES)}~\cite{EU-Regulation-EES-InternalDocument-2017} that is planned to be an automated IT system for registering third-country nationals. But also in the context of forensic investigation, face data is processed and exchanged between member states to identify suspected criminals or find missing individuals - implemented as part of the \textit{Schengen Information System (SIS)}\cite{EU-Regulation-SIS-1862-2018}. Despite the widespread need for secure and convenient FR systems, their performance is still heavily influenced by the quality of the compared images - a well-known challenge that has been extensively studied by researchers, standardisers, and practitioners alike. 

\subsection{Face Image Quality}

In order to avoid inaccurate results in cooperative image capturing scenarios, \textit{quality assessment algorithms} are employed to eliminate low-quality samples prior to being processed by the recognition system. In this context, \textit{ISO/IEC CD3 29794-5} \cite{ISO-IEC-29794-5-CD3-FaceQuality-231018} distinguishes between \textit{unified} and \textit{component quality measures} represented as integer values in the range of $[0, 100]$, with a value of $100$ indicating perfect value to a receiving face recognition system (\textit{perfect utility}~\cite{ISO-IEC-29794-1-FDIS-FaceQuality-231013}). More precisely, unified quality scores are designed to predict whether a given sample negatively or positively contributes to the recognition accuracy, thus aggregating the qualities of all variation factors in a given face image. While unified quality scores are suitable for reducing the workload~\cite{OsorioRoig-QualityScoresIndexing-IWBF-2022} and enhancing the performance~\cite{schlett2022face} of a biometric system, their aggregated nature makes it impossible to determine the factors responsible for causing the low quality. Further, current state-of-the-art methods for estimating unified quality scores~\cite{meng2021magface}\cite{hernandez2020biometric}\cite{terhorst2020ser} rely on deep neural networks that are known to trade explainability for accuracy gains caused by complex model architectures. 

\begin{figure}
\centering
\includegraphics[width=0.5\linewidth]{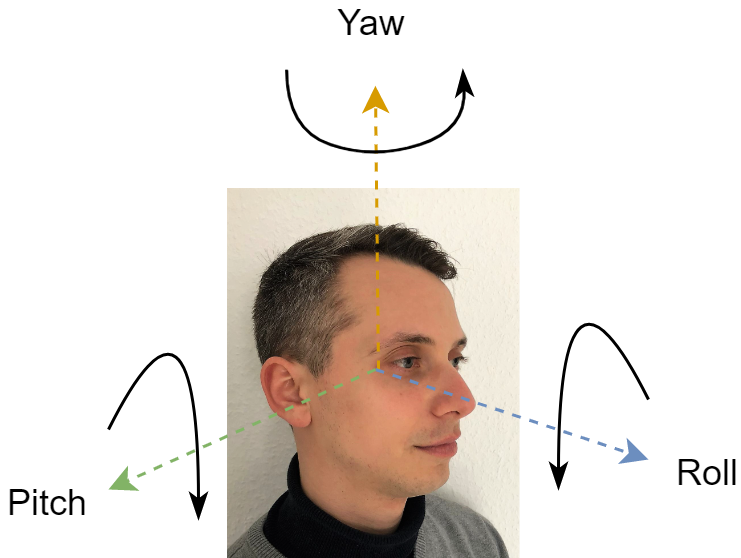}
\caption{Each facial pose is represented by a yaw, pitch, and roll angle as specified by ISO/IEC 39794-5~\cite{ISO-IEC-39794-5-G3-FaceImage-191015}.\label{fig:intro-yaw-pitch-roll}}
\end{figure}

In contrast, component quality measures are developed to improve explainability by breaking down the quality of each variation factor expressed by an individual component scalar. Consequently, operators of FR systems can provide actionable feedback, asking subjects to correct their behavior concerning factors such as head frontalisation or expression neutrality. In practice, one of the main challenges in developing \textit{component quality assessment algorithms} is to disentangle all facial attributes to analyse each of them in isolation. To overcome this obstacle, we exploit in this work the fine-grained controllability of synthetically generated face images to derive a lightweight pose quality assessment estimator.       

\subsection{3D-Aware Face Image Generation}

Traditionally, the generation of synthetic face images with \textit{2D-aware generative adversarial networks} (GANs) \cite{karras2019style} \cite{karras2020analyzing} \cite{karras2021alias} yields a remarkable degree of photorealism that is indistinguishable from real data by human perception \cite{nightingale2022ai}\cite{shen2021study}. After generating synthetic face images, their facial attributes must be edited to analyse how this type of variation affects the similarity scores.

\begin{figure}
\centering
\setlength{\tabcolsep}{1pt}
\resizebox{0.6\linewidth}{!}{%
\begin{tabular}{ccc}
\includegraphics[width=.3\linewidth,valign=m, height=.12\linewidth]{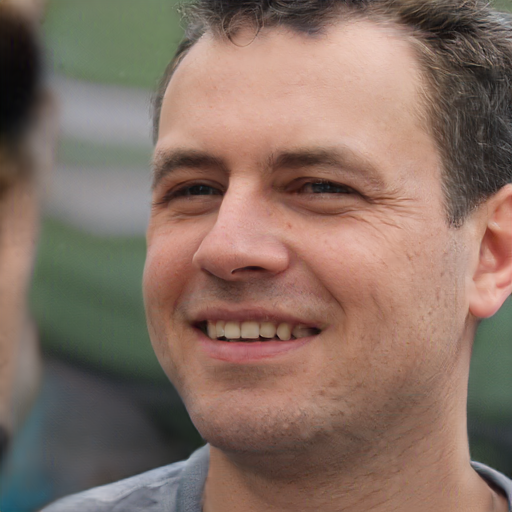} & \includegraphics[width=.3\linewidth,valign=m, height=.12\linewidth]{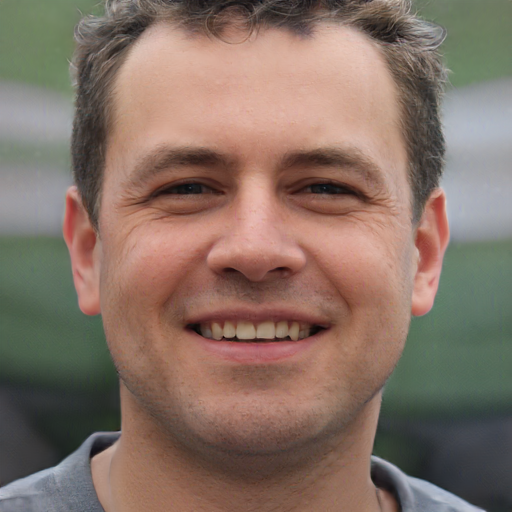} &
\includegraphics[width=.3\linewidth,valign=m, height=.12\linewidth]{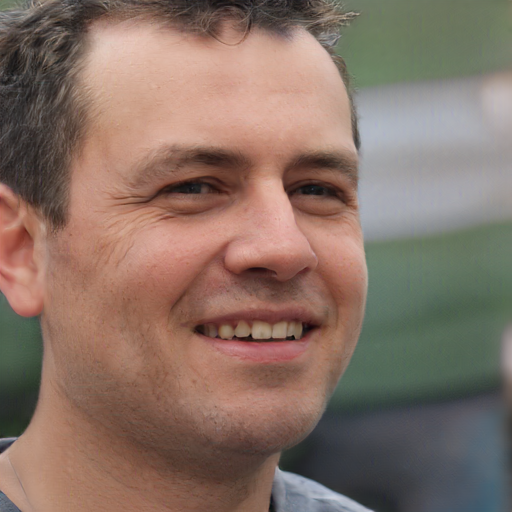} \\

\includegraphics[width=.3\linewidth,valign=m, height=.12\linewidth]{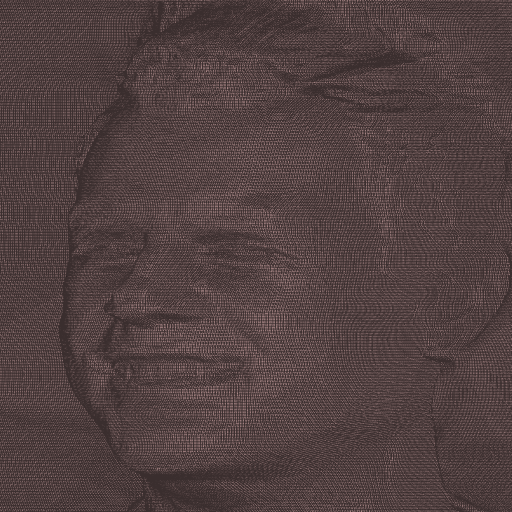} & \includegraphics[width=.3\linewidth,valign=m, height=.12\linewidth]{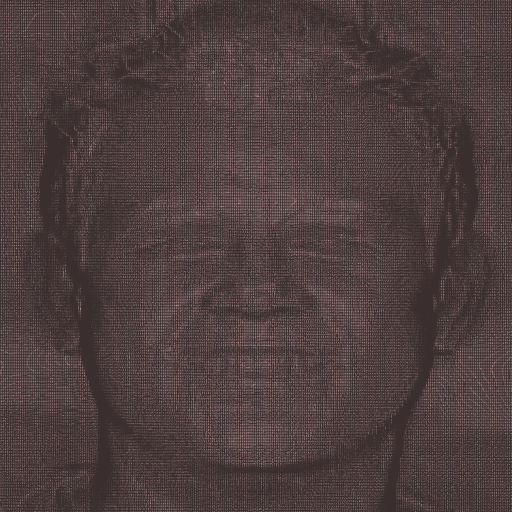} & \includegraphics[width=.3\linewidth,valign=m, height=.12\linewidth]{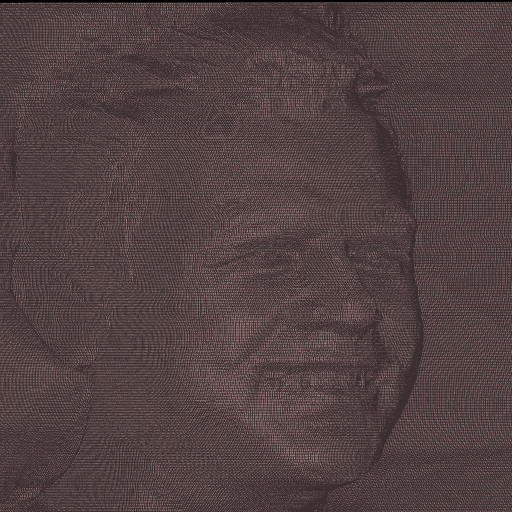}

\end{tabular}}
\caption{3D-aware face images and their corresponding 3D-meshes generated with EG3D \cite{chan2022efficient}, demonstrating the high visual quality of StyleGAN2~\cite{karras2020analyzing} combined with consistent face geometry.\label{fig:intro-3d-examples}}
\end{figure}

Despite the capabilities of 2D-aware GANs in producing highly photorealistic images, they often have difficulty preserving the facial geometry when manipulating attributes such as yaw and pitch angle (see Figure~\ref{fig:intro-yaw-pitch-roll}), thus resulting in loss of crucial identity information. Therefore, a new branch of research constitutes the development of \textit{3D-aware GANs} to enable the precise editing of facial attributes while preserving identity, in particular consistency in facial geometry~\cite{chan2022efficient}. Compared to previous works~\cite{nguyen2019hologan}\cite{niemeyer2021giraffe}\cite{medin2022most}, EG3D addresses the trade-off between geometry consistency and memory utilization by incorporating an explicit-implicit network architecture (see Section\ref{sec:related-works}). In this work, we leverage the 3D awareness of EG3D as shown in Figure~\ref{fig:intro-3d-examples} to investigate the influence of yaw and pitch on different FR systems based on synthetic data. \textcolor{red}{Specifically, all face images are generated with an EG3D generator pre-trained on a pose-balanced version of the FFHQ\cite{karras2019style, chan2022efficient} dataset with a resolution of $512^2$.}
    
\subsection{Contribution and Paper Structure}

The generation of our synthetic dataset is first described in Section \ref{sec:dbs}, complemented by a 3D consistency check using an off-the-shelf \textit{structure-from-motion} \cite{schoenberger2016sfm} and \textit{Multi-View Stereo}\cite{schoenberger2016mvs} pipeline. This 3D consistency check is mandatory as editing pitch and yaw angles with GANs typically corresponds with changing other facial characteristics caused by their entanglement in the latent space~\cite{shen2020interfacegan}. Next, Section~\ref{sec:estimator} introduces our proposed \textit{SYP-Lasso} pose quality estimator together with the current implementation of pose component quality as defined by \textit{ISO/IEC CD3 29794-5}~\cite{ISO-IEC-29794-5-CD3-FaceQuality-231018}. In Section \ref{sec:pose-impact-estimation}, we analyse the effect of various yaw-pitch combinations on the biometric performance of five FR systems: ArcFace~\cite{deng2019arcface}, MagFace~\cite{meng2021magface}, CurricularFace~\cite{huang2020curricularface}, AdaFace~\cite{kim2022adaface}, and Cognitec~\footnote{\url{https://www.cognitec.com/} (FaceVACS, Version 9.6.0)}~\footnote{\textbf{Disclaimer:} Please be advised that the evaluations conducted using the Cognitec SDK in this work were performed solely by the authors. The results presented do not necessarily reflect the full capabilities of the technology}. Finally, Section \ref{sec:pose-quality-estimation} demonstrates the effectiveness of our SYP-Lasso regression model on the \textit{BIWI Kinect Face dataset}~\cite{fanelli2011real} as it outperforms state-of-the-art component and unified quality assessment algorithms on the task of pose-dependent quality prediction.  

Overall, the contributions of this work can be summarised as follows:

\begin{itemize}
    \item Proposal of a 3D-aware synthetic dataset (\textit{Syn-YawPitch}), comprised of 1,000 identities with a total of 144,000 images, covering various yaw-pitch angle combinations.
    \item Estimation of the impact of different yaw-pitch combinations based on state-of-the-art open source and COTS FR systems.
    \item Proposal of a lightweight ISO/IEC-compliant~\cite{ISO-IEC-29794-5-CD3-FaceQuality-231018} pose quality estimation algorithm~\cite{tibshirani1996regression} (\textit{SYP-Lasso}) that yields explainable quality measures.
\end{itemize}

\section{Related Work}
\label{sec:related-works}

This section first summarises the recent advancements in 2D-aware face image synthesis, followed by introducing works focusing on the transition to 3D-aware face image generation. Finally, Section \ref{sec:synthetic-data-in-biometics} describes how synthetic data was used in the past to evaluate FR systems.

\subsection{2D-Aware Face Image Synthesis}

Analysing synthetic data in the context of biometric applications requires image synthesis models to generate facial images with their quality indistinguishable from real data. That to say, Karras et al.~\cite{karras2017progressive} were among the first to propose a GAN architecture able to generate high-quality images (1024x1024) achieved by progressively growing the image resolution processed by the generator and discriminator. With StyleGAN \cite{karras2019style}, a style-based GAN architecture was introduced to avoid the entanglement of features observed in the latent space of traditional GANs. The new concept of an intermediate latent space, such as the possibility to mix styles of multiple identities, paved the way for modern unconditional GANs to gain additional control over their outputs.

With StyleGAN reaching new milestones in unconditioned image synthesis and latent space disentanglement, its usage is still restricted by noticeable image artefacts caused by the integration of adaptive instance normalization~\cite{huang2017arbitrary} operations and major flaws in the architecture network design~\cite{karras2020analyzing}. With the follow-up works of StyleGAN2 \cite{karras2020analyzing} and StyleGAN3 \cite{karras2021alias}, the typical droplet-shaped artefacts and texture-sticking effects of StyleGAN were eliminated, yielding face images indistinguishable from real ones by the human eye \cite{shen2021study}\cite{nightingale2022ai}.

\subsection{3D-Aware Face Image Synthesis}

Although traditional GANs achieve highly realistic face images, they still struggle with learning fundamental 3D structures necessary to manipulate facial attributes, such as poses or expressions \cite{nguyen2019hologan}. To overcome this obstacle, another branch of development heads towards geometry-aware GANs \cite{nguyen2019hologan}\cite{niemeyer2021giraffe}\cite{medin2022most}\cite{chan2022efficient}, learning 3D representations from 2D training data. Typically, 3D object representations can be learned either explicitly, e.g., through deep voxel grids \cite{sitzmann2019deepvoxels}, or implicitly through latent representations \cite{mildenhall2021nerf}. The explicit processing of voxel grids through 3D convolutions is very computationally extensive, limiting the practical usage on a large scale for high-resolution images. In contrast, while implicit architectures are more compact, they are often too slow due to large fully connected neural networks required to store all information, often restricting their usage to single-image optimization scenarios.

To improve computation efficiency and memory utilization, Chan et al.~\cite{chan2022efficient} proposed an explicit-implicit approach, combining the advantages of both worlds. They introduce Tri-plane 3D representations based on a StyleGAN2 \cite{karras2020analyzing} generator backbone to represent spatial information through neural radiance fields explicitly \cite{mildenhall2021nerf}. Subsequently, these features are passed to a Tri-plane decoder, which implicitly learns more compact representations of the underlying faces. Finally, a differentiable neural volume renderer transforms the given features into a 2D facial image. \textcolor{red}{To decouple head pose angles from other facial attributes, the StyleGAN2 mapping network is conditioned with camera parameters~\cite{hempel20226d} that are randomly swapped with a probability of 50\% during training. When generating face images during inference, this conditioning mechanism has shown to significantly reduce the entanglement of head poses to other variation factors observed in the training dataset, such as smiling expressions for frontal portraits.} In this work, we leverage these advancements in 3D-aware image synthesis to generate a large-scale database ("Syn-YawPitch") with consistent yaw and pitch angle variations to derive pose quality estimators as required for \textit{ISO/IEC CD3 29794-5}~\cite{ISO-IEC-29794-5-CD3-FaceQuality-231018}.

\subsection{Synthetic Data in Biometrics}
\label{sec:synthetic-data-in-biometics}

The recent advances in computer vision and computer graphics towards generating realistic and controllable face images led to various applications benefiting biometric systems. As shown by Joshi et al.~\cite{joshi2022synthetic}, synthetic datasets are widely used for various applications in the context of human analysis, including the evaluation of biometric performances, assessing the scalability of biometric systems, and reducing biases by increasing the size and diversity of real datasets. Using synthetic data either to replace or enrich real data grounds on the high degree of realism and controllability synthetic samples provide - a trend we expect to continue in the future. Among the research using synthetic data in FR, one of the pioneering works of Kortylewski et al.~\cite{kortylewski2019analyzing} uses 3DMM face models to analyse the generalization capability of FR systems during inference by training on synthetic datasets with held-out pose intervals.

Colbois et al.~\cite{colbois2021use} create synthetic face images containing variations similar to those of the Multi-PIE dataset~\cite{gross2010multi} to compare the biometric performance across six FR systems to the performance reported on the original Multi-PIE dataset. By observing only minor performance differences, the authors conclude that their synthetic dataset can be used as a substitute of real data to evaluate the performance of FR systems, while complying to privacy protection. A similar conclusion was drawn by Zhang et al.~\cite{zhang2021applicability} who analysed the differences between synthetic and real face images quantitatively using three quality assessment algorithms. In a follow-up work, Grimmer et al.~\cite{grimmer2022time} used two face age progression techniques~\cite{alaluf2021only}\cite{shen2020interfacegan} to generate mated samples with age differences of up to $60$ years to investigate how much the facial identity changes over time.

Previously, Grimmer et al.~\cite{grimmer2021generation} generate non-deterministic mated samples by exploring new directions in the latent space of StyleGAN by applying Principle Component Analysis. Creating mated samples in a non-deterministic manner aims to detect vulnerabilities in FR systems that would stay undetected when focusing on simple facial attributes (\textit{e.g., smiling}) only.

\section{Syn-YawPitch}
\label{sec:dbs}

The derivation of a pose quality estimator (PQE) requires the isolated analysis of yaw and pitch angle combinations to assess their impact on FR accuracy. Hence, each covariate must be precisely controlled while leaving other facial characteristics and environmental factors fixed. In this work, we leverage advances in 3D-aware image generation by using the work of Chan et al.~\cite{chan2022efficient} (\textit{EG3D}). Specifically, we generate a database comprised of synthetic mated samples with varying combinations of yaw and pitch angles (\textit{Syn-YawPitch)}.

Although roll angles can affect the biometric performance, this type of variation can be reduced by applying face alignment algorithms. As such, all EG3D training samples were preprocessed, thus the generator cannot manipulate roll angles during inference. However, we point to the findings of Lu et al.~\cite{lu2019experimental}, indicating stable biometric performances for roll angles in the range of $\pm30^{\circ}$. We further demonstrate in Section~\ref{sec:SYP-Lasso-vs-baseline} that the majority of face images in our evaluation database have roll angles within $\pm20^{\circ}$.                   

We generate facial images of $1,000$ identities (IDs) to evaluate the biometric performance for different yaw-pitch combinations. Specifically, we construct a grid-like pairing of yaw ($\Phi_y$) and pitch ($\Phi_p$) angles defined by the Cartesian product of $\Phi_y=\{90\degree, 80\degree, 60\degree, 40\degree, 20\degree, 0\degree\}$ and $\Phi_p=\{90\degree, 80\degree, 60\degree, 40\degree, 20\degree, 0\degree\}$. Following this approach, the total number of images in the Syn-YawPitch dataset rises to $144,000$, comprising $144$ images per identity. A thorough comparison of Syn-YawPitch to existing datasets that comprise head pose variations is outlined in Table~\ref{tab:head-pose-dbs}.

\begin{table}[]
\centering
\caption{Overview of publicly available datasets with head pose variations. Notably, Syn-YawPitch stands out among the constrained datasets with the highest number of images per subject, including a grid-like distribution of yaw-pitch combinations to facilitate systematic evaluation protocols.}
\label{tab:head-pose-dbs}
\resizebox{0.99\linewidth}{!}{%
\begin{tabular}{|l|l|l|l|l|l|}
\hline
\textbf{Dataset} & \textbf{Subjects} & \textbf{Images} & \textbf{Synthetic?} & \textbf{Constrained?} & \textbf{Pose Variation} \\ \hline
Syn-YawPitch (ours)     & 1,000             & 144,000         & Yes                 & Yes                   & Yaw, Pitch              \\ \hline
CMU Multi-PIE~\cite{gross2010multi}   & 337               & $\sim$750,000   & No                  & Yes                   & Yaw                     \\ \hline
SynHead~\cite{gu2017dynamic}          & 10                & $\sim$510,000   & Yes                 & Yes                   & Yaw, Pitch, Roll        \\ \hline
SASE~\cite{lusi2016sase}             & 50                & 30,000          & No                  & Yes                   & Yaw, Pitch, Roll        \\ \hline
Pointing'04~\cite{gourier2004estimating}      & 15                & 2,940           & No                  & Yes                   & Yaw, Pitch              \\ \hline
AFLW-2000~\cite{zhu2016face}        & 2,000             & 2,000           & No                  & No                    & Yaw, Pitch, Roll        \\ \hline
BIWI~\cite{fanelli2011real}             & 20                & $\sim$15,000    & No                  & Partly                & Yaw, Pitch, Roll        \\ \hline
CAS-PEAL-R1~\cite{4404053}      & 1,040             & $\sim$30,000    & No                  & Yes                   & Yaw, Pitch              \\ \hline
FEI~\cite{thomaz2010new}              & 200               & 2,800           & No                  & Yes                   & Yaw, Pitch              \\ \hline
M2FPA~\cite{li2019m2fpa}            & 229               & $\sim$400,000   & No                  & Yes                   & Yaw, Pitch              \\ \hline

%Colour FERET~\cite{phillips2000feret}     & 994               & 11,338          & No                  & Yes                   & Yaw                     \\ \hline

CPLFW~\cite{CPLFWTech}     & 5,749               & 13,233          & No                  & No                   & In-the-wild                     \\ \hline

\end{tabular}}
\end{table}

\begin{figure}
\centering
\setlength{\tabcolsep}{1pt}
\begin{tabular}{ccccc}
 \mbox{} &  $-34^{\circ}$ & $-24^{\circ}$ & $24^{\circ}$ & $45^{\circ}$ \\
$-21^{\circ}$ & \includegraphics[width=.2\linewidth,valign=m]{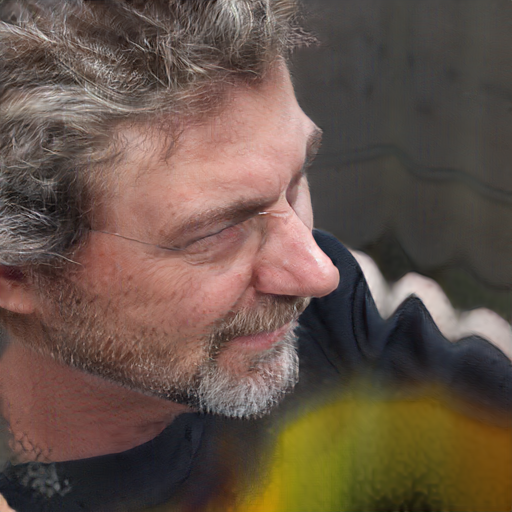} & \includegraphics[width=.2\linewidth,valign=m]{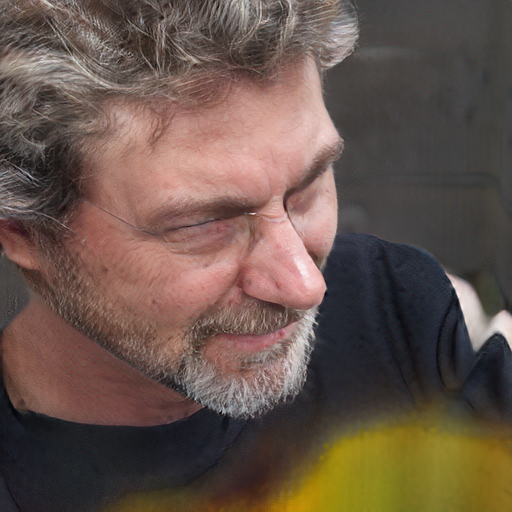} & \includegraphics[width=.2\linewidth,valign=m]{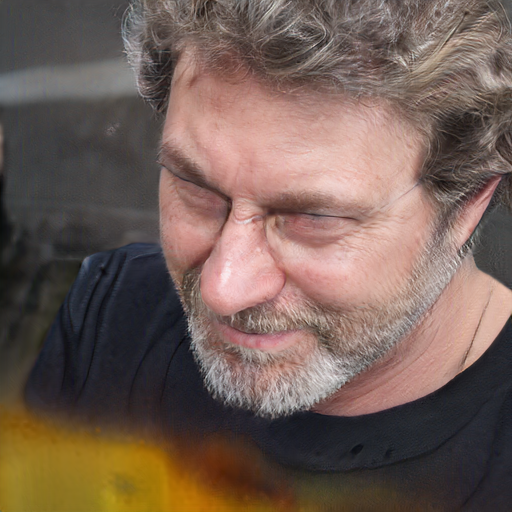} & \includegraphics[width=.2\linewidth,valign=m]{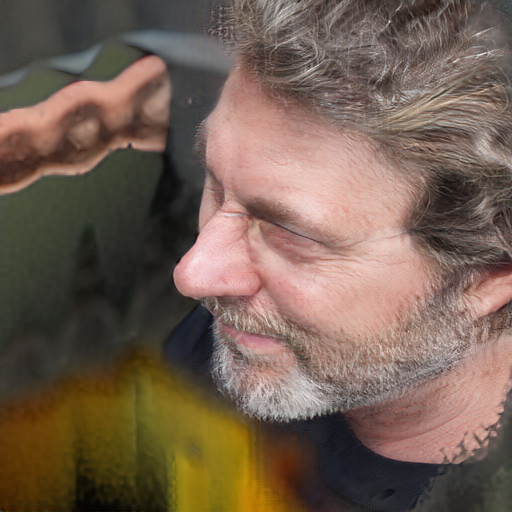} \\

$-14^{\circ}$ & \includegraphics[width=.2\linewidth,valign=m]{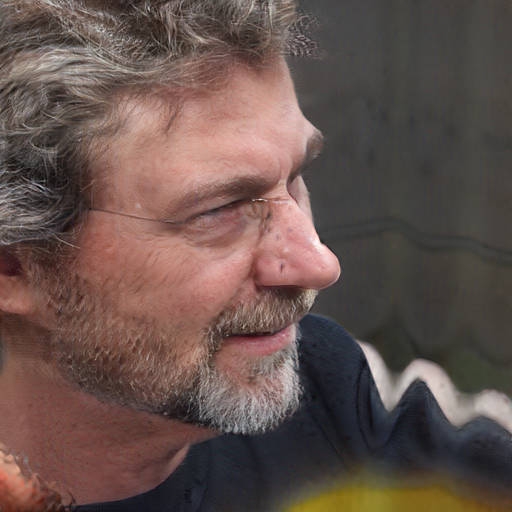} & \includegraphics[width=.2\linewidth,valign=m]{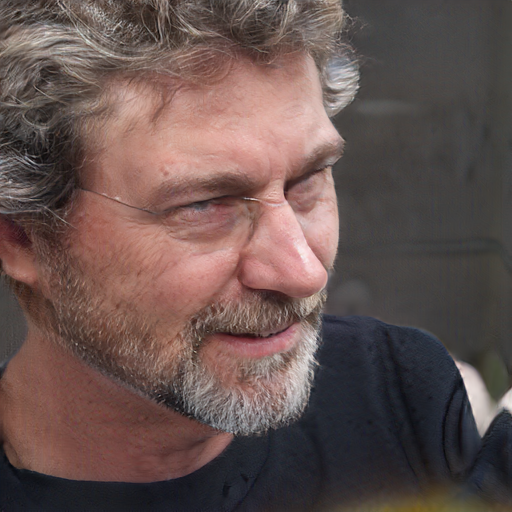} & \includegraphics[width=.2\linewidth,valign=m]{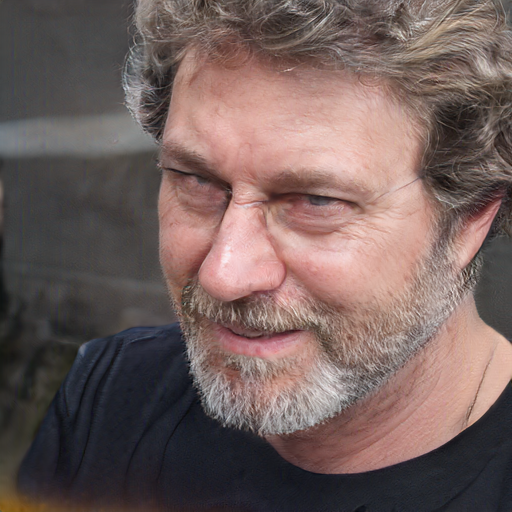} & \includegraphics[width=.2\linewidth,valign=m]{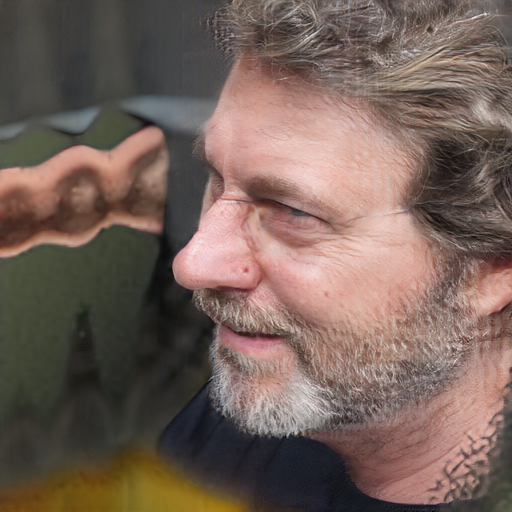}\\

$12^{\circ}$ & \includegraphics[width=.2\linewidth,valign=m]{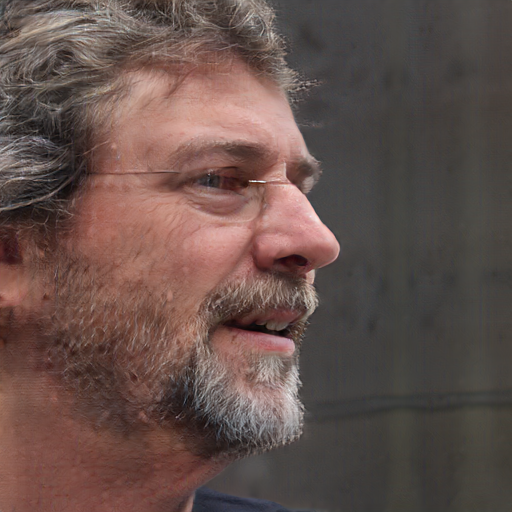} & \includegraphics[width=.2\linewidth,valign=m]{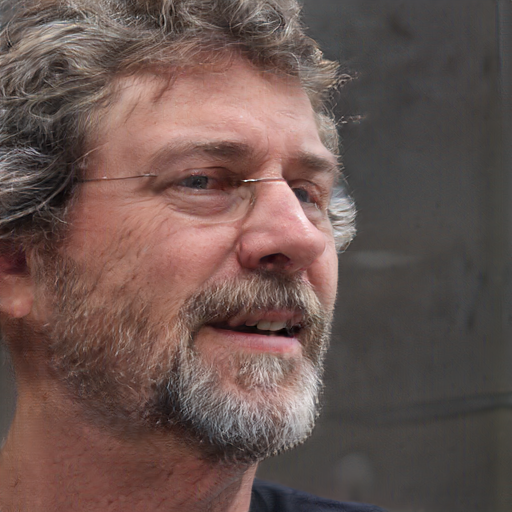} & \includegraphics[width=.2\linewidth,valign=m]{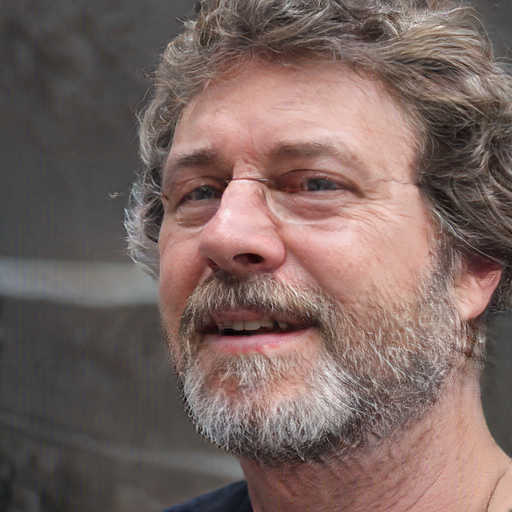} & \includegraphics[width=.2\linewidth,valign=m]{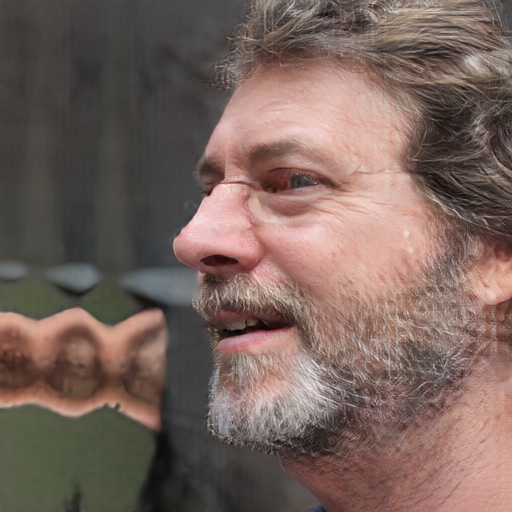}\\

$34^{\circ}$ & \includegraphics[width=.2\linewidth,valign=m]{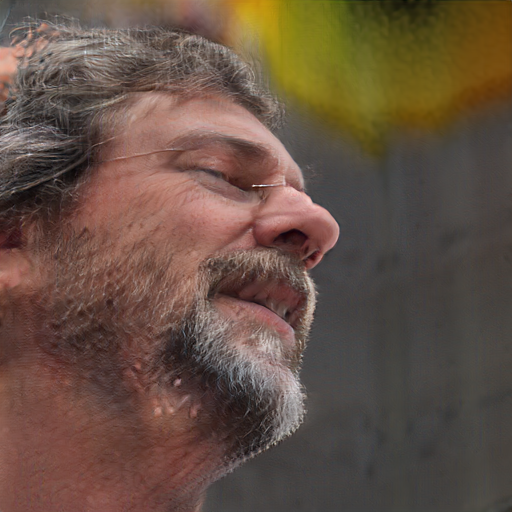} & \includegraphics[width=.2\linewidth,valign=m]{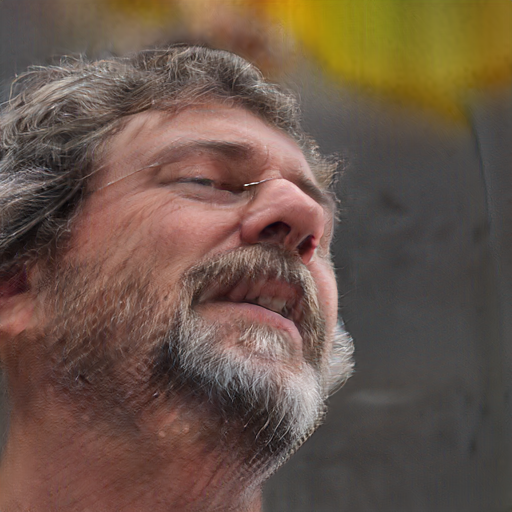} & \includegraphics[width=.2\linewidth,valign=m]{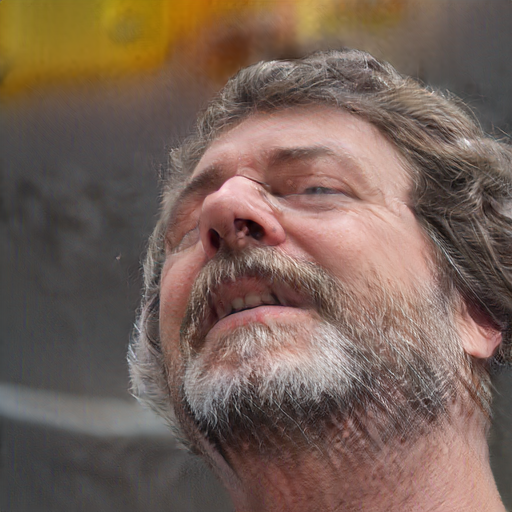} & \includegraphics[width=.2\linewidth,valign=m]{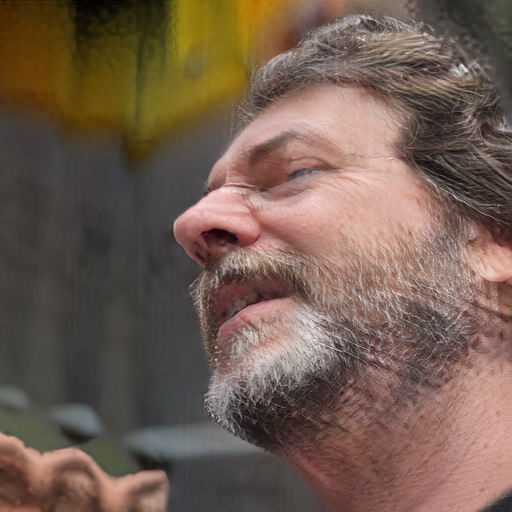}\\
\end{tabular}
\caption{Examples of Syn-YawPitch samples illustrating the visual quality of several yaw-pitch angle combinations. These examples highlight the ability of EG3D\cite{chan2022efficient} to change the camera angles solely without inducing biases observed in frontal views, such as smiling facial expressions. \label{fig:example-imgs}}
\end{figure}

\subsection{Head Pose Re-Annotation}
\label{sec:syn-yawpitch-reannotation}

Although EG3D allows the editing of yaw and pitch angles with consistent face geometry, we identified discrepancies between the specified and actual angles. This effect amplifies the more the specified angles deviate from the frontal viewing angle and can likely be traced back to the head pose imbalances within the training dataset of EG3D. However, an accurate estimation of the actual head pose angles is a crucial precondition to estimating their impact on FR utility. Therefore, we re-annotate each sample with an off-the-shelf pose estimator \cite{hempel20226d}. More specifically, we define a function that maps the original EG3D pose labels $\Phi_y$ and $\Phi_p$ to the median value obtained by applying the pose estimator on 100 randomly generated images $\{x_1, x_2, \dots, x_{100}\}$: 

\begin{equation}
\begin{aligned}
\hat \Phi_y &= \text{Median}( \mathcal{P}_{\Phi_y}(x_1), \mathcal{P}_{\Phi_y}(x_2), \dots, \mathcal{P}_{\Phi_y}(x_{100}))
\end{aligned}
\end{equation}
and 
\begin{equation}
\begin{aligned}
\hat \Phi_p &= \text{Median}( \mathcal{P}_{\Phi_p}(x_1), \mathcal{P}_{\Phi_p}(x_2), \dots, \mathcal{P}_{\Phi_p}(x_{100}))
\end{aligned}
\end{equation}

with $\hat{\Phi}_y, \hat{\Phi}_p$ denoting the median yaw and pitch values, $\mathcal{P}_{\Phi_y}(x_i), \mathcal{P}_{\Phi_p}(x_i)$ referring to the estimated yaw and pitch angles of the i-th randomly generated face image $x_{i}$. Once all median values are computed, the resulting lookup table translates EG3D pose labels to more accurate ones based on the external head pose estimator \cite{hempel20226d}. Example images of the re-annotated samples are shown in Figure~\ref{fig:example-imgs}. Note that through re-annotating our Syn-YawPitch database, we identify imbalances in the generation capability of EG3D - \textit{e.g.}, specifying $\Phi_y=90^{\circ}$ corresponds to an adjusted value of $\hat{\Phi_y}=45^{\circ}$, whereas flipping the sign to $\Phi_y=-90^{\circ}$ translates to $\hat{\Phi_y}=-34^{\circ}$. We reason these pose imbalances are adopted during the training stage of EG3D caused by less represented pose intervals within the training dataset (\textit{FFHQ}\cite{karras2019style}. Despite the ability of EG3D to generate pose angles beyond those seen in the training phase, we discourage the selection of angles above $90^{\circ}$ as the number of artefacts increases visibly.

\begin{figure}
\centering
\setlength{\tabcolsep}{1pt}
\resizebox{0.99\linewidth}{!}{%
\begin{tabular}{lccccc}
 \mbox{} &  Frontal & $\phi_p=-21^{\circ}$ & $\phi_p=34^{\circ}$ & $\phi_y=-34^{\circ}$ & $\phi_y=45^{\circ}$ \\
EG3D & \includegraphics[width=.2\linewidth,valign=m]{{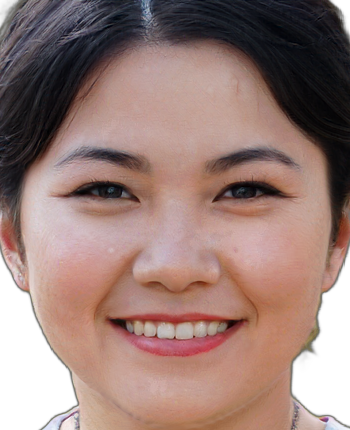}} & \includegraphics[width=.2\linewidth,valign=m]{{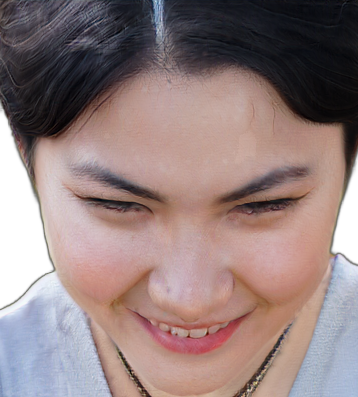}} & \includegraphics[width=.2\linewidth,valign=m]{{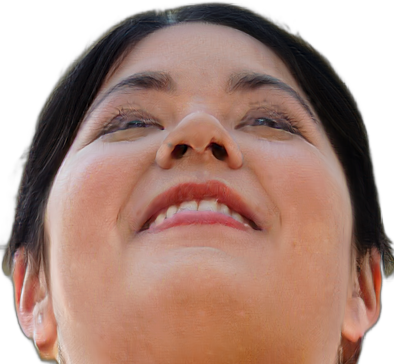}} & \includegraphics[width=.2\linewidth,valign=m]{{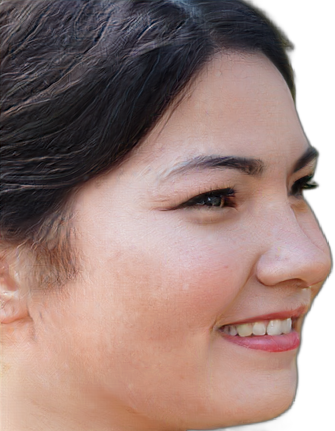}} & \includegraphics[width=.2\linewidth,valign=m]{{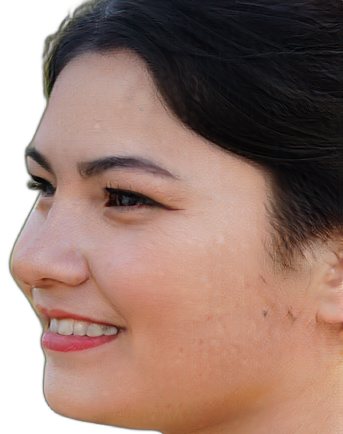}} \\

FLAME & \includegraphics[width=.2\linewidth,valign=m]{{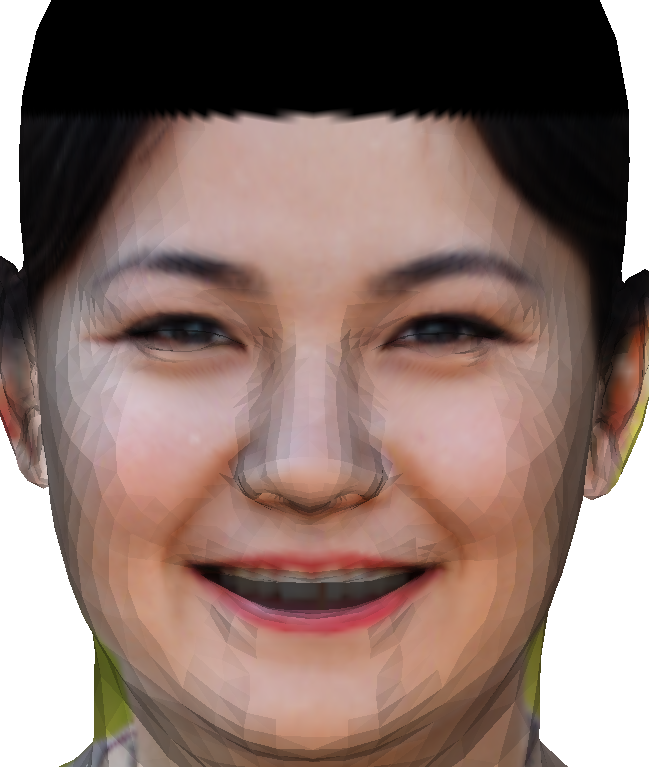}} & \includegraphics[width=.2\linewidth,valign=m]{{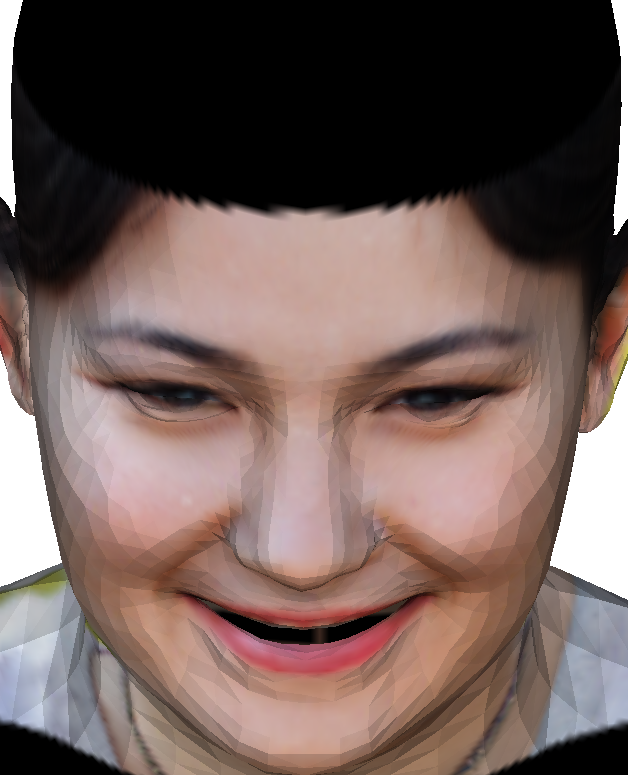}} & \includegraphics[width=.2\linewidth,valign=m]{{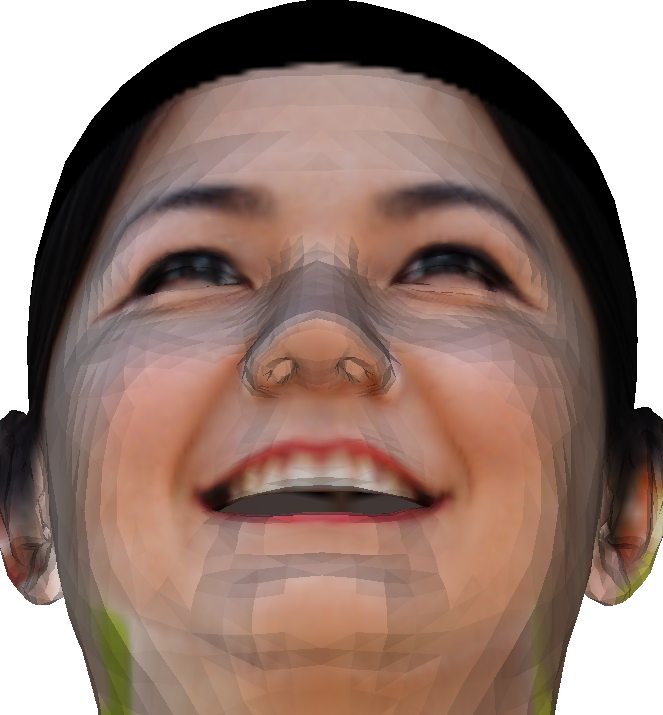}} & \includegraphics[width=.2\linewidth,valign=m]{{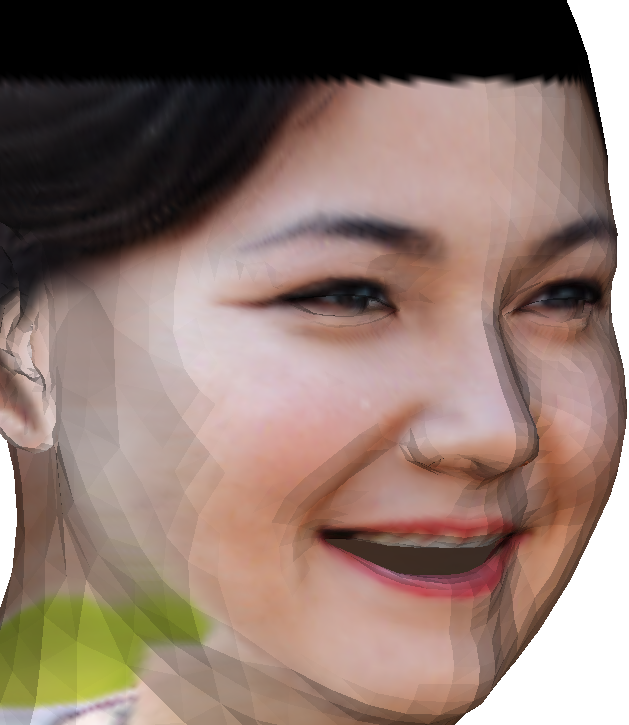}} & \includegraphics[width=.2\linewidth,valign=m]{{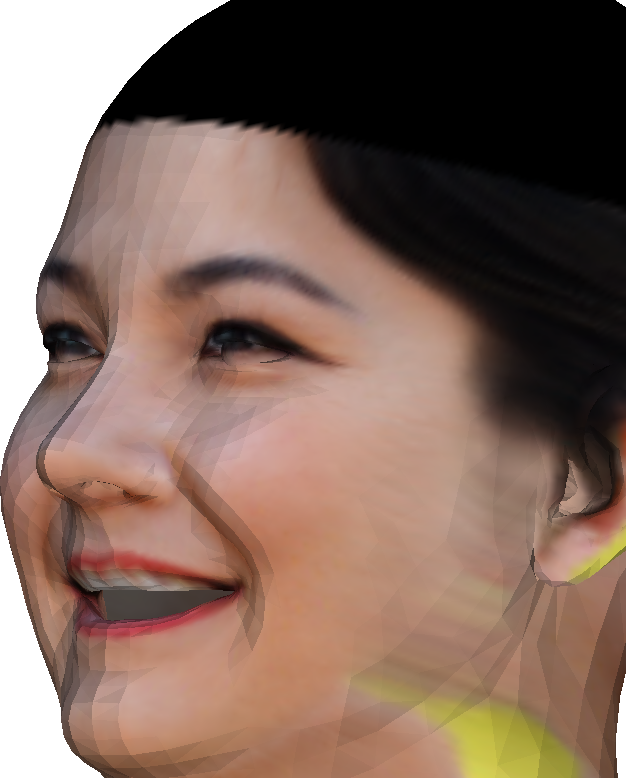}} \\
\end{tabular}}
\caption{\textcolor{red}{Visual comparison between face images generated with a 3D-aware GAN (EG3D~\cite{chan2022efficient}) and a 3D Morphable Face Model (FLAME~\cite{li2017learning}).\label{fig:flame-vs-eg3d-comp}}}
\end{figure}

\textcolor{red}{In Figure~\ref{fig:flame-vs-eg3d-comp}, we provide an example illustrating the visual quality of EG3D and a state-of-the-art \textit{3D Morphable Face Model} (3DMM) \cite{li2017learning, danvevcek2022emoca} from distinct pose angles. While the parametrization of 3DMMs enables precise control over diverse facial attributes, the rendered face images lack photorealism and fail to capture important details, such as facial hair, teeth, or eyeballs.}

\subsection{Implementation Details}
\label{sec:implement-details}

Since the architecture of EG3D adopts the generator of StyleGAN2 \cite{karras2020analyzing}, we follow the recommendation of Colbois et al.~\cite{colbois2021use} by restricting each latent code to keep a minimum distance to each previously drawn sample. As shown by \cite{colbois2021use}, this sampling strategy anticipates unnaturally high lookalike rates caused by latent vectors drawn within close regions in the latent space. 

For the generation of Syn-YawPitch, we choose a truncation factor of $0.5$ to generate facial images with stable image quality. The truncation factor can be interpreted as a parameter that controls the trade-off between perceived image quality and diversity~\cite{zhang2021applicability} by constraining the latent space region from which latent vectors are randomly drawn. We further exploit the high flexibility of synthetic data to construct a separate dataset to compute representative non-mated comparison scores using a truncation factor of $0.75$. This choice is inspired by Zhang et al.~\cite{zhang2021applicability}, who have demonstrated that truncation factors above $0.75$ yield similar non-mated comparison score distributions than those observed with real datasets.

\subsection{Geometry-Consistency}

\textcolor{red}{To ensure that EG3D\cite{chan2022efficient} preserves the facial geometry when manipulating viewing angles, we apply an independent Structure-from-Motion (SfM)~\cite{schoenberger2016sfm} technique used to reconstruct three-dimensional structures from two-dimensional image sequences. Hence, given a set of 2D face images from multiple viewpoints, SfM aims to estimate the underlying 3D structure. In this section, we examine the ID preservation of an example face image in terms of facial geometry based on a 3D-aware (EG3D~\cite{chan2022efficient}) and 2D-aware GAN (StyleGAN~\cite{karras2019style}).}

\begin{figure}
\centering
\setlength{\tabcolsep}{1pt}
\resizebox{0.90\linewidth}{!}{%
\begin{tabular}{cccc}
\includegraphics[width=0.23\linewidth,valign=m]{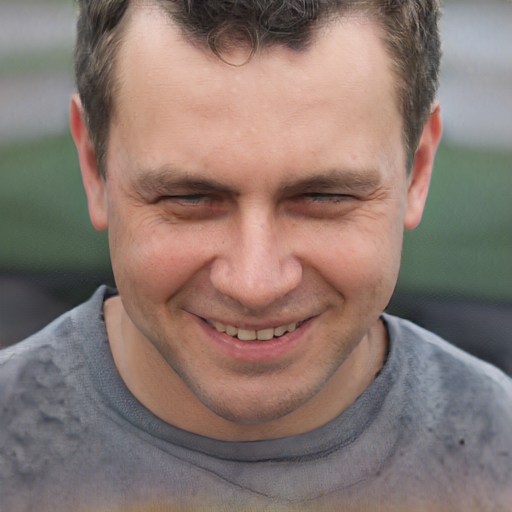} & \includegraphics[width=0.23\linewidth,valign=m]{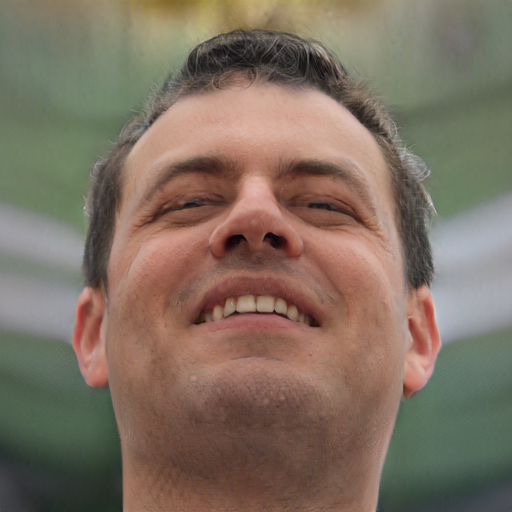} & \includegraphics[width=0.23\linewidth,valign=m]{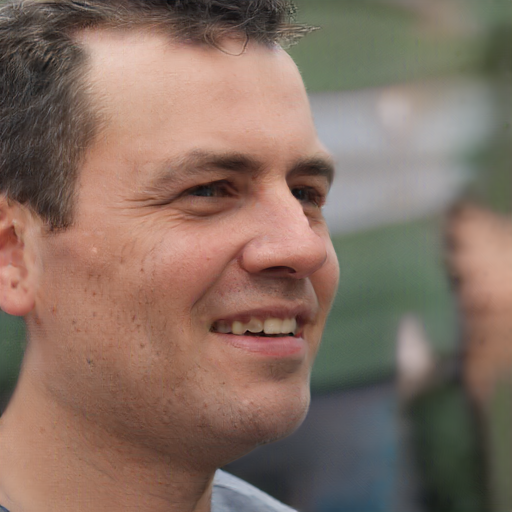} & \includegraphics[width=0.23\linewidth,valign=m]{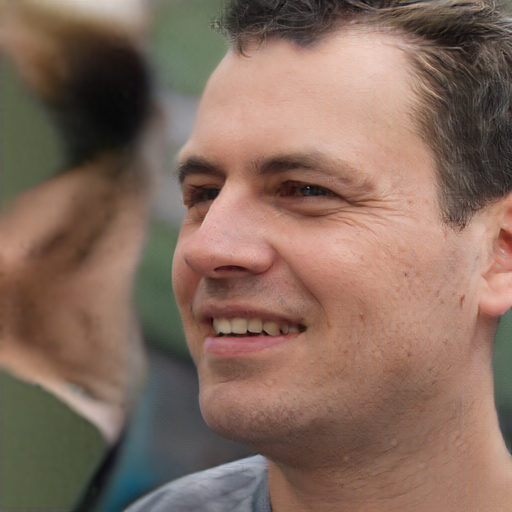} \\

\includegraphics[width=0.23\linewidth,valign=m]{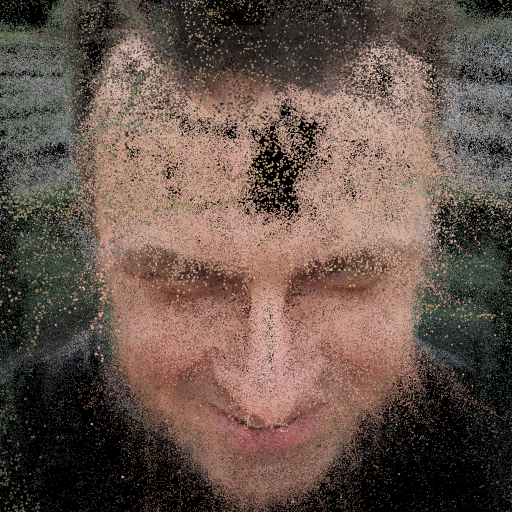} & \includegraphics[width=0.23\linewidth,valign=m]{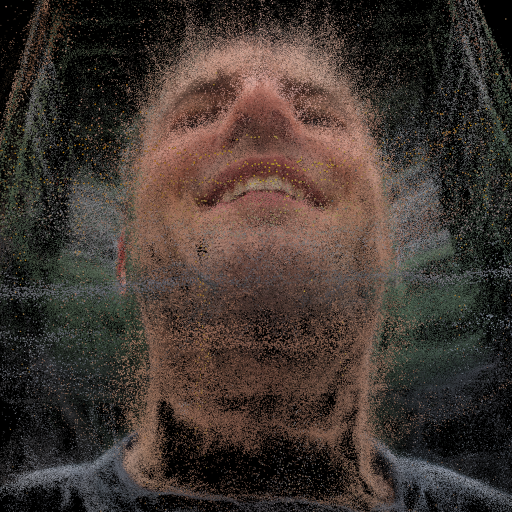} & \includegraphics[width=0.23\linewidth,valign=m]{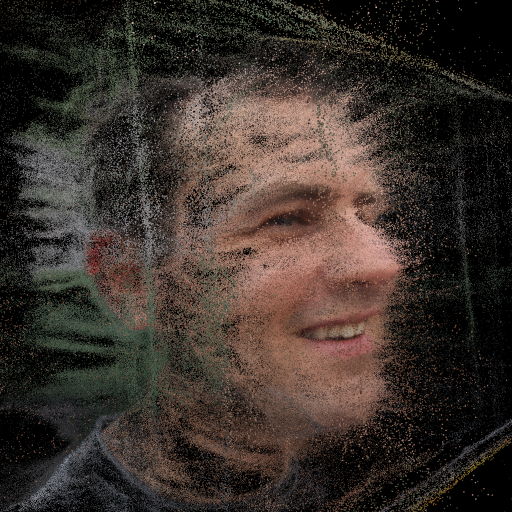} & \includegraphics[width=0.23\linewidth,valign=m]{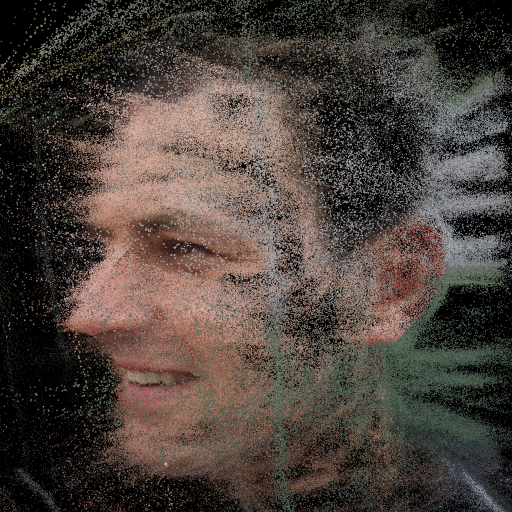}\\

\mbox{} & \mbox{} & \mbox{} & \mbox{} \\

\includegraphics[width=0.23\linewidth,valign=m]{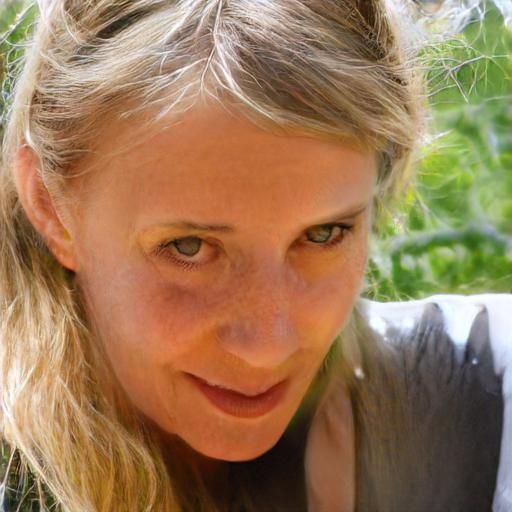} & \includegraphics[width=0.23\linewidth,valign=m]{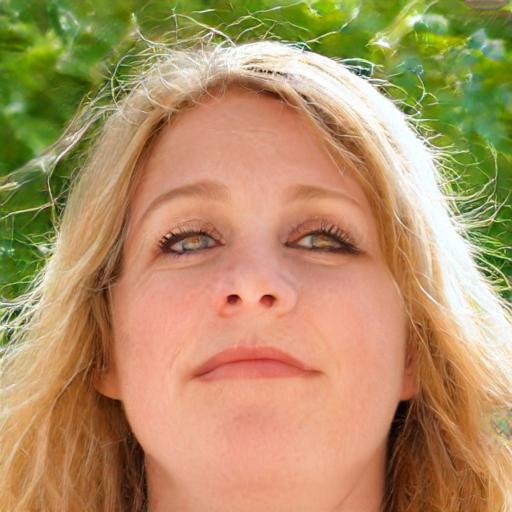} & \includegraphics[width=0.23\linewidth,valign=m]{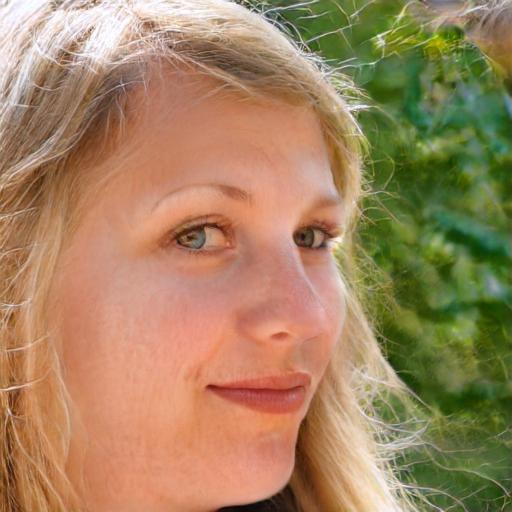} & \includegraphics[width=0.23\linewidth,valign=m]{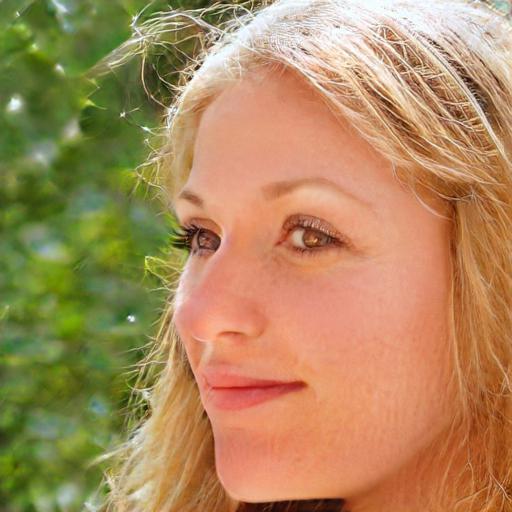} \\

\includegraphics[width=0.23\linewidth,valign=m]{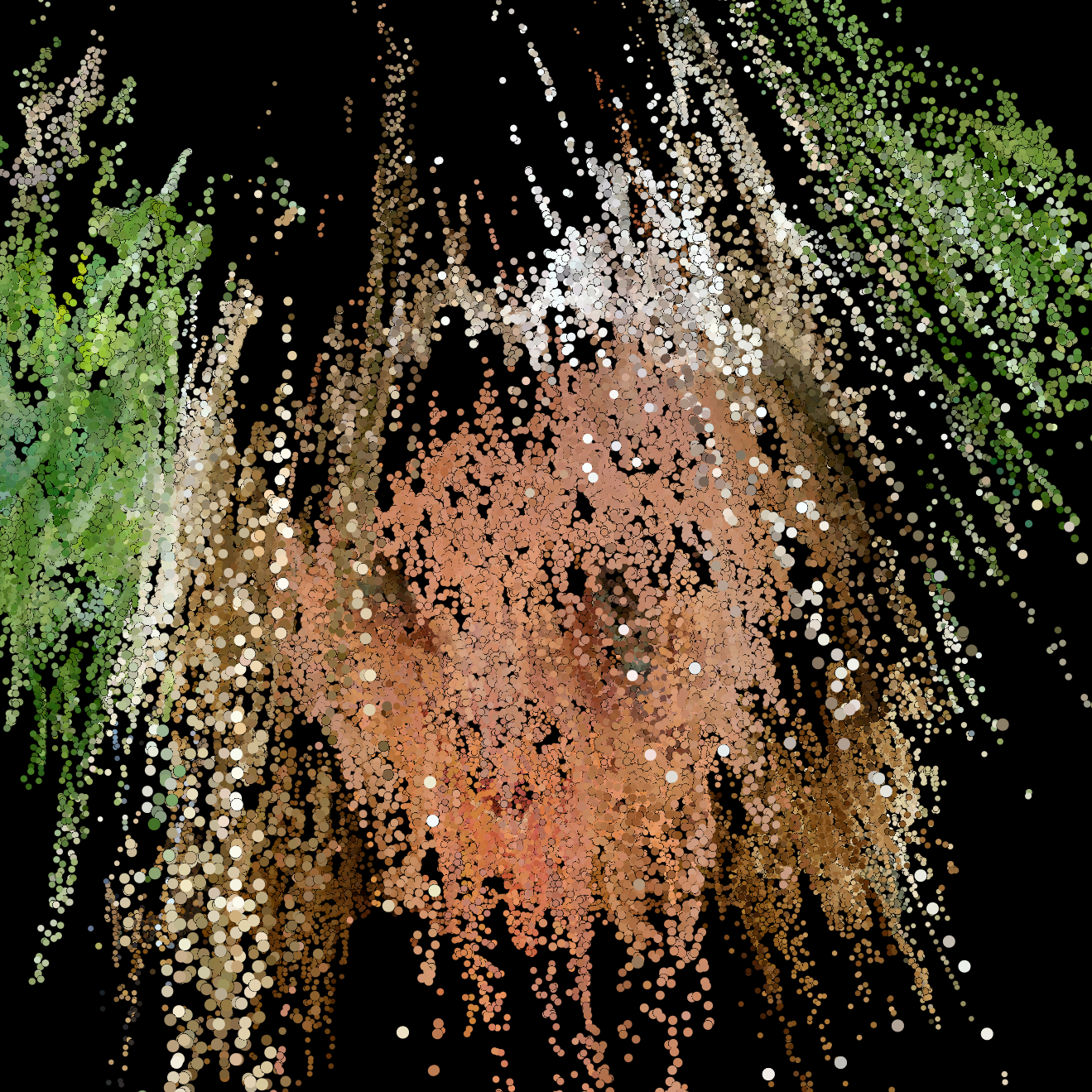} & \includegraphics[width=0.23\linewidth,valign=m]{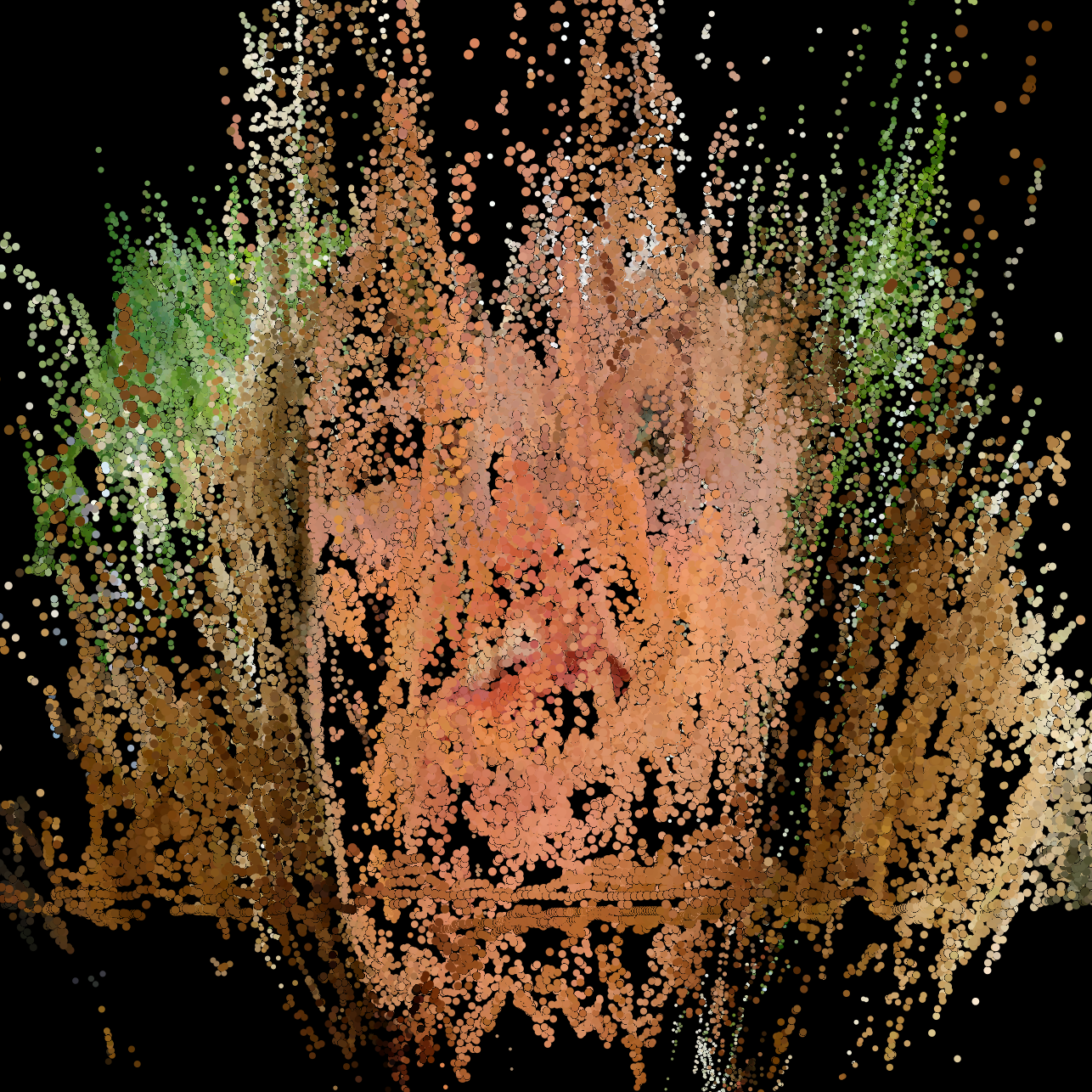} & \includegraphics[width=0.23\linewidth,valign=m]{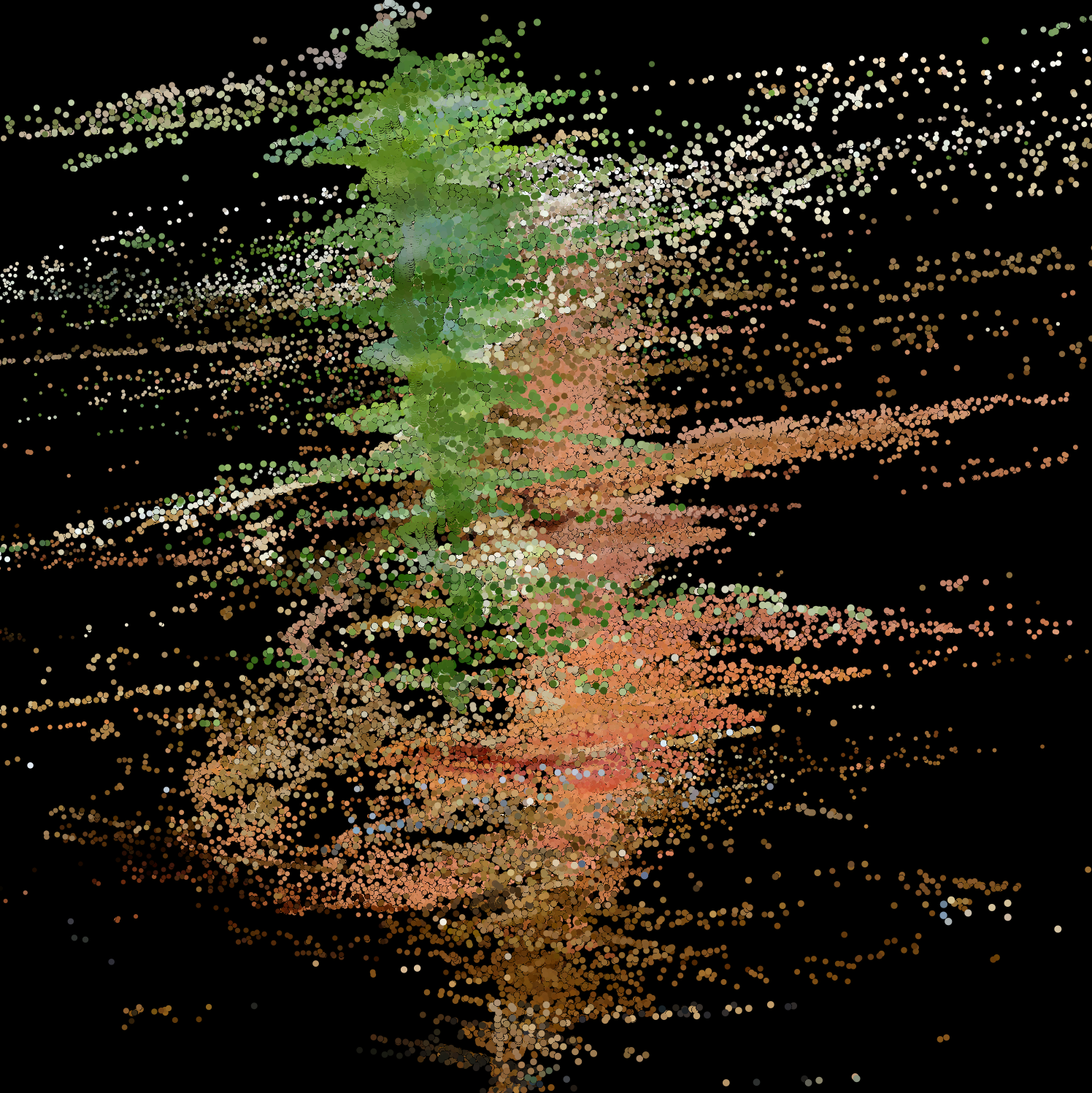} & \includegraphics[width=0.23\linewidth,valign=m]{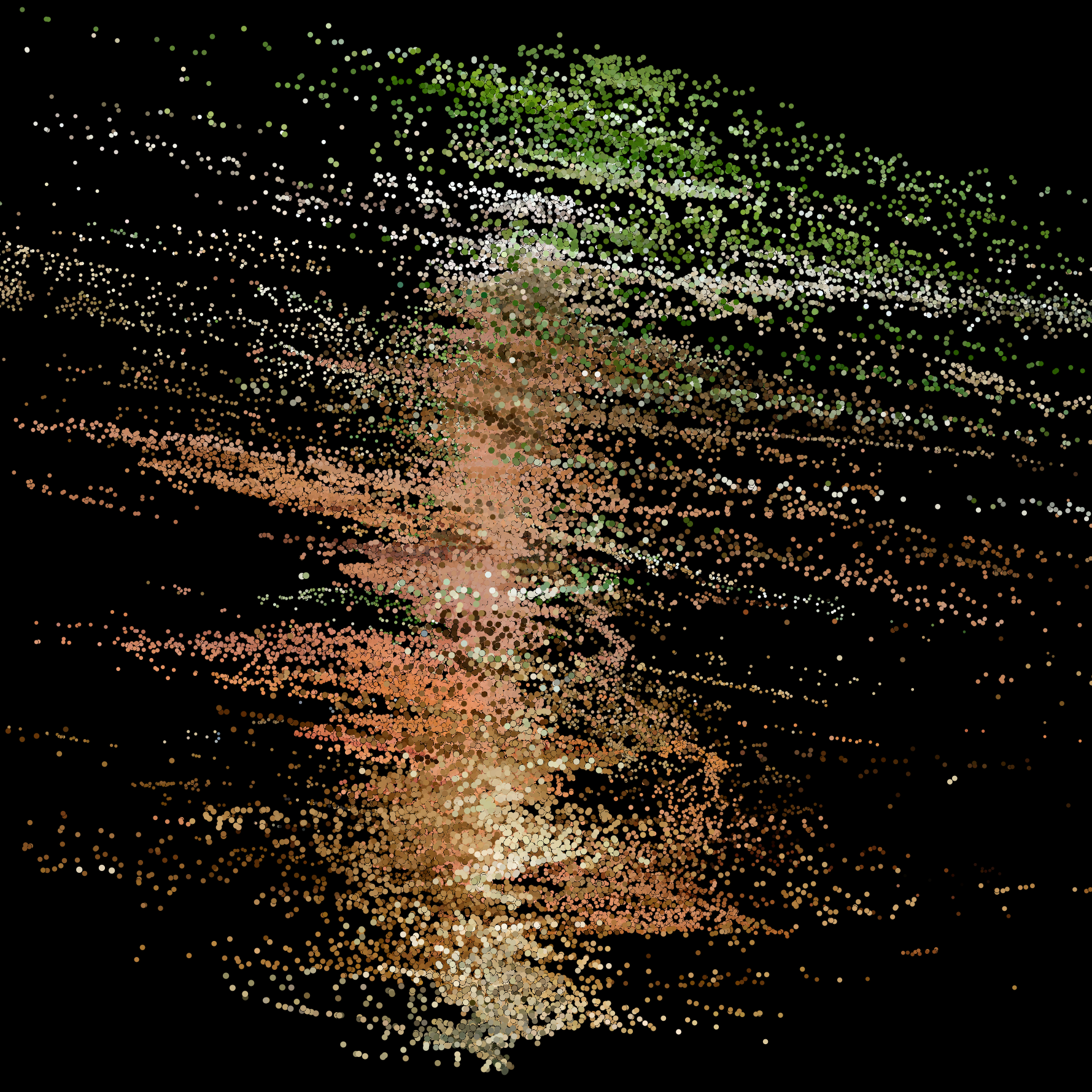} \\
\end{tabular}}
\caption[Caption for LOF]{Example face images generated with EG3D \cite{chan2022efficient} (Top) and InterFaceGAN \cite{shen2020interfacegan} (Bottom). The 3D point clouds were generated using an open source \textit{Structure-from-Motion}~\cite{schoenberger2016sfm} software tool.\protect\footnotemark\label{fig:geom-consistency}. \textcolor{red}{Given a sequence of EG3D face images of the same subject from multiple viewpoints, the resulting dense point cloud reconstructions show a coherent face structure. In contrast, the 2D-aware InterFaceGAN method~\cite{shen2020interfacegan} fails to generate geometry-consistent image sequences, as indicated by non-existing patterns in the 3D point clouds.}}
\end{figure}

We observe that, despite the high image quality achieved by StyleGAN, its generator is not optimized to preserve facial geometry when changing the camera view. A demonstration of this issue is provided in Figure \ref{fig:geom-consistency}, illustrating the 3D point cloud reconstruction~\cite{schoenberger2016sfm} derived from $320$ facial images of the same synthetic ID generated from various viewing angles. Given an initial latent code, InterFaceGAN \cite{shen2020interfacegan} is used to generate mated samples by manipulating yaw and pitch angles, respectively, using tiny-sized steps in the latent space. The point clouds indicate that StyleGAN, by default, is unsuitable for preserving facial geometry when editing yaw or pitch angles, as depicted in Figure \ref{fig:geom-consistency}. 

In contrast, the 3D-aware network architecture of EG3D includes a neural volume rendering module conditioned on the image's camera extrinsics and intrinsics. With this setup applied during training, EG3D can subsequently be used for inference with consistent control over yaw and pitch angles. Example images are visualised in Figure \ref{fig:geom-consistency}, showing the facial images of the same ID simulated from multiple viewing angles.In contrast to InterFaceGAN, it is visible that EG3D preserves the geometric face structure much better, utilising the same amount of $320$ input images with varying yaw-pitch angles. With the precondition of consistent face geometry being fulfilled, we have established a solid foundation for the experimental results presented in the next sections.   

\footnotetext{\url{https://colmap.github.io/}}

\section{Pose Quality Estimator}
\label{sec:estimator}

This section introduces the PQE derived from our Syn-YawPitch dataset, aiming to predict the FR utility of a face image given its yaw and pitch angles. In addition, we introduce the current implementation of ISO/IEC CD3 29794-5~\cite{ISO-IEC-29794-5-CD3-FaceQuality-231018} to serve as a baseline PQE in our experiments. Note that our proposed PQE, namely \textit{SYP-Lasso}, and the ISO/IEC-based PQE are implemented to fulfil the standardization requirements for quantifying the utility of a given face image, constraining the quality measure to be an integer scalar in the range of $[0, 100]$, where 0 indicates the lower and 100 represents the upper utility boundaries.

\subsection{Baseline PQE}
\label{sec:baseline-predictors}

To establish a fair benchmark, we utilize the component quality measures defined by ISO/IEC CD3 29794-5~\cite{ISO-IEC-29794-5-CD3-FaceQuality-231018} for estimating utility based on head pose angles for yaw and pitch, respectively:

\begin{equation}
    \text{ISO-PQE}(\Phi_y) = 100\cdot cos^2\Phi_y
    \label{eq:iso-pqe-yaw}
\end{equation}

and 

\begin{equation}
    \text{ISO-PQE}(\Phi_p) = 100\cdot cos^2\Phi_p
    \label{eq:iso-pqe-pitch}
\end{equation}

Contrary to this work, the utility estimations defined in ISO/IEC CD3 29794-5\cite{ISO-IEC-29794-5-CD3-FaceQuality-231018} are computed separately for yaw and pitch angles, which neglects the potential inter-correlations between both factors. To improve the comparability to our proposed SYP-Lasso model, we choose the minimum of $\Phi_y$ and $\Phi_p$ to derive a fused pose component quality measure.

\subsection{SYP-Lasso PQE}

\begin{table}
\centering
\caption{\textcolor{red}{An overview of the coefficients of our SYP-Lasso PQE trained on Syn-YawPi using 5 different face recognition systems. Each coefficient can be interpreted as the contribution of the corresponding pose angle to the expected similarity when compared to frontal face images. Note that in this table, we use a polynomial degree of $n=1$ to facilitate an easy interpretation of the coefficients.\label{tab:lasso-model-coefs}}}
\resizebox{0.99\linewidth}{!}{%
\begin{tabular}{l c c c c c}
\hline 
\vspace{0.005cm}
FR System & $\phi_{p[-]}$ & $\phi_{p[+]}$ & $\phi_{y[-]}$ & $\phi_{y[+]}$  \\
\hline
\vspace{0.005cm}
ArcFace~\cite{deng2019arcface} & $1.12\mathrm{e}{-02}$ & $-1.13\mathrm{e}{-02}$ & $8.73\mathrm{e}{-03}$ & $-6.27\mathrm{e}{-03}$ \\

Magface~\cite{meng2021magface} & $1.14\mathrm{e}{-02}$ & $-1.17\mathrm{e}{-02}$ & $8.76\mathrm{e}{-03}$ & $-6.17\mathrm{e}{-03}$ \\

CurricularFace~\cite{huang2020curricularface} & $1.18\mathrm{e}{-02}$ & $-1.23\mathrm{e}{-02}$ & $9.20\mathrm{e}{-03}$ & $-6.68\mathrm{e}{-03}$ \\

AdaFace~\cite{kim2022adaface} & $1.23\mathrm{e}{-02}$ & $-1.26\mathrm{e}{-02}$ & $1.02\mathrm{e}{-02}$ & $-7.06\mathrm{e}{-03}$ \\

Cognitec\footnotemark[1] & $8.83\mathrm{e}{-03}$ & $-1.07\mathrm{e}{-02}$ & $6.67\mathrm{e}{-03}$ & $-4.70\mathrm{e}{-03}$ \\

\hline
\end{tabular}
}
\label{table:financial_results}
\end{table}

Given the similarity scores as a result of comparing frontal face images to images of the same identity but with modified yaw and pitch angles, it is possible to capture the relationship between head pose and similarity score with a mathematical model. Thereby, we make the following assumptions to derive our proposed SYP-Lasso PQE: 1) When comparing a frontal reference to a head-rotated probe sample, we assume that a low similarity score is directly caused by the induced variation in yaw and pitch angles. This statement is grounded on the fact that only the camera viewing angle is changed, thus causing variation in yaw and pitch while leaving other facial attributes fixed. 2) Since the similarity scores depend only on a single factor of variation, we further assume that similarity can be interpreted as a utility measure that complies with the requirements defined in ISO/IEC CD3 29794-5~\cite{ISO-IEC-29794-5-CD3-FaceQuality-231018}.

\textcolor{red}{Given our proposed Syn-YawPitch dataset with its grid-like distribution of yaw-pitch combinations, we compare frontal face images to their mated counterparts under different yaw-pitch constellations, modelling the relationship between similarity score (\textit{i.e.} pose quality measure) versus head pose angles using a lasso regression model~\cite{tibshirani1996regression} that minimizes the following term:}

\begin{equation}
\label{eq:lasso-opt}
\underset{\beta\in\mathbb{R}^{T}}{\min} \left\{ \frac{1}{N}||s - \mathcal{P}\beta||_{2}^2 +\lambda||\beta||_{1} \right\}
\end{equation}

\textcolor{red}{In Equation \ref{eq:lasso-opt}, $s = [s_{1}, s_{2}, \dots, s_{N}]$ represents the similarity scores for mated comparisons of the pose-rotated probe samples and the frontal reference images, where $N=144,000$ denotes the size of the Syn-YawPitch dataset. Furthermore, $\beta = [\beta_{1}, \beta_{2},\dots, \beta_{T}]$ denotes the lasso regression coefficients computed by minimizing Equation~\ref{eq:lasso-opt}\footnote{\url{https://scikit-learn.org/stable/modules/generated/sklearn.linear_model.Lasso.html}} with an empirically chosen regularization term  $\lambda=1\mathrm{e}{-06}$ to reduce overfitting and perform feature selection by automatically removing covariates with less predictive value. Finally, each row in $\mathcal{P} \in \mathbb{R}^{N \times T}$ contains the $T=\left|\mathcal{C}\right|$ regression covariates described as:}

\begin{equation}
\color{red}
\label{eq:lasso-covariates}
\begin{split}
\mathcal{C} = &\bigcup_{i=1}^{n} \Bigl\{\phi_{p[+]}^{i}, \phi_{p[-]}^{i}, \phi_{y[+]}^{i}, \phi_{y[-]}^{i}\Bigr\} \\
&\mbox{} \bigcup \  \Bigl\{ a\cdot b \ | \ (a, b) \in \Phi_{p}^{i-1} \times \Phi_{y}^{i-1}  \Bigr\}
\end{split}
\end{equation}

\textcolor{red}{with a polynomial degree $n$, pitch angles $\Phi_{p} = \{\phi_{p[+]}, \phi_{p[-]}\}$ and yaw angles $\Phi_{y} = \{\phi_{y[+]}, \phi_{y[-]}\}$ separated into their positive and negative components to enable SYP-Lasso to learn pose-related performance imbalances essential for deriving a pose utility measure. }

\begin{figure}
\label{fig:adjR-vs-polyDegree}
\centering
\includegraphics[width=0.6\linewidth]{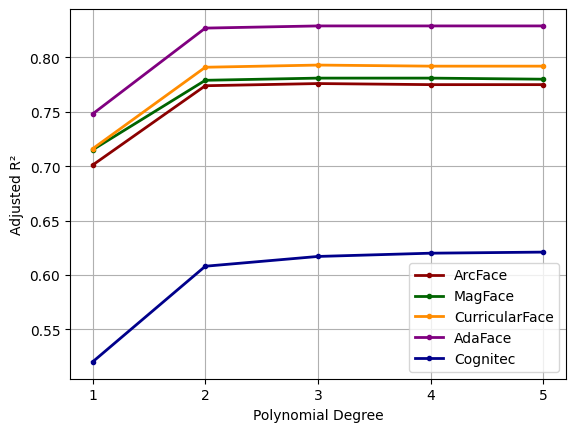}
\caption{The curves illustrate only minor gains in the Adjusted $R^2$ for polynomial degrees of $n > 2$. Note that in Section~\ref{sec:SYP-Lasso-vs-baseline} we choose a polynomial degree of $n=2$ due to the best trade-off between model complexity and performance in pose quality estimation.    \label{fig:poly-vs-adjR}}
\end{figure}

\textcolor{red}{As shown in Figure~\ref{fig:poly-vs-adjR}, only a minor performance gain is achieved by increasing the polynomial degree beyond $n=2$ based on the adjusted $R^2$ metric that expresses the percentage of variance in the target variable (\textit{i.e., similarity score}) that can be explained by the covariates. Given the interpretability of the coefficients when selecting $n=1$, Table~\ref{tab:lasso-model-coefs} first depicts the raw impact of each covariate, with a more thorough discussion following in Section~\ref{sec:pose-impact-estimation}. Ultimately, our proposed SYP-Lasso model is configured with $n=2$ as it yields the best trade-off between model complexity and utility estimation performance, reaching an adjusted $R^2$ of up to $83\%$ for the open source FR systems and $61\%$ for FaceVACS (Cognitec), as seen in Figure~\ref{fig:poly-vs-adjR}.}        

\textcolor{red}{Ultimately, during inference, SYP-Lasso simplifies to the setup of a simple linear regression. Additionally, we scale and clip the estimated pose utility to the range of $[0, 100]$ to fulfil the requirements of \textit{ISO/IEC CD3 29794-5}}.

%\begin{equation}
%\label{eq:lasso-eq}
%\text{SYP-Lasso}(\mathcal{P}) =
%\begin{cases} 
%      0 & , \ \mathcal{P}\beta\leq 0 \\
%      100 \cdot \mathcal{P}\beta & , \ 0 \leq \mathcal{P}\beta\leq 1 \\
%      100 & , \ 1 \leq \mathcal{P}\beta 
%\end{cases}
%\end{equation}

% Synth vs Real samples

\section{Pose Impact Estimation}
\label{sec:pose-impact-estimation}

The generation of Syn-YawPitch allows the impact analysis of different yaw-pitch angle combinations on the biometric recognition performance. We exploit the controlled structure of our synthetic dataset to compute \textit{equal error rate} (EER), \textit{false non-match rate} (FNMR), and \textit{false match rate} (FMR) to assess the performance of the FR systems included in our experiments. Specifically, we report FNMRs and FMRs at fixed 1\% FMR (i.e., FNMR@1\%FMR) and FNMR (i.e., FMR@1\%FNMR) errors to evaluate the biometric performance.  

\begin{figure*}
\label{fig:geometry-consistency}

\centering
\resizebox{0.7\linewidth}{!}{%
  \begin{subfigure}[b]{\linewidth}
    \includegraphics[width=\linewidth]{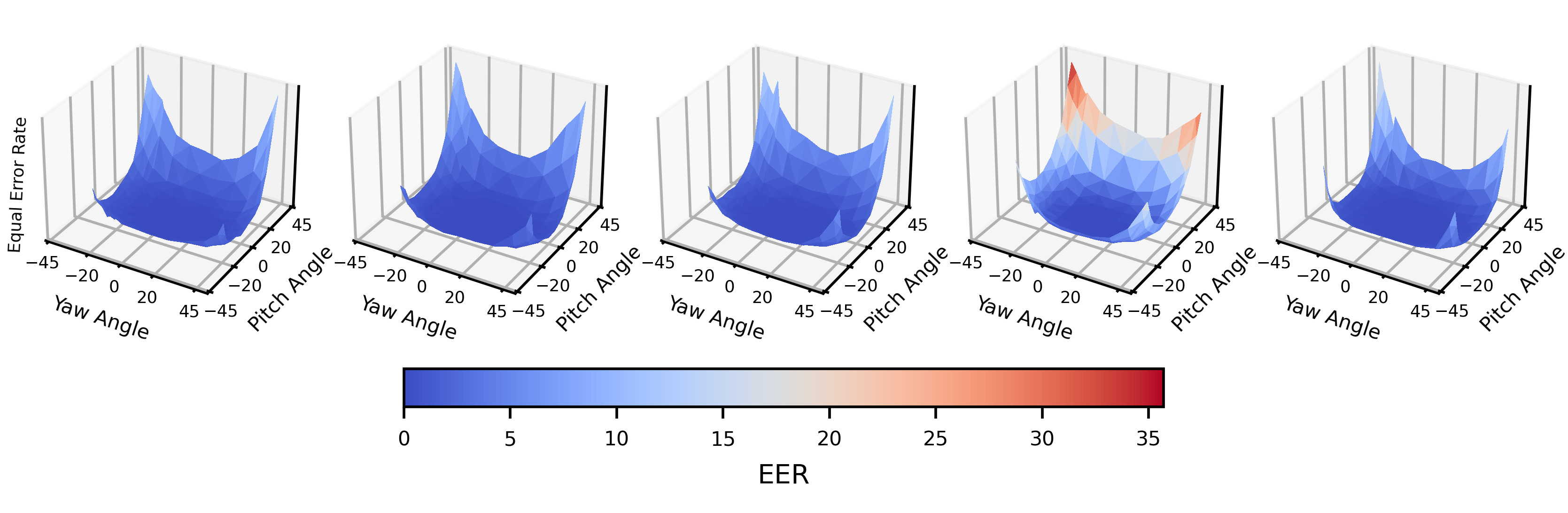}
    %\caption{StyleGAN2 \cite{karras2020analyzing}}
    \label{fig:synth-face-example}
  \end{subfigure}}
  
 \resizebox{0.7\linewidth}{!}{%
  \begin{subfigure}[b]{\linewidth}
    \includegraphics[width=\linewidth]{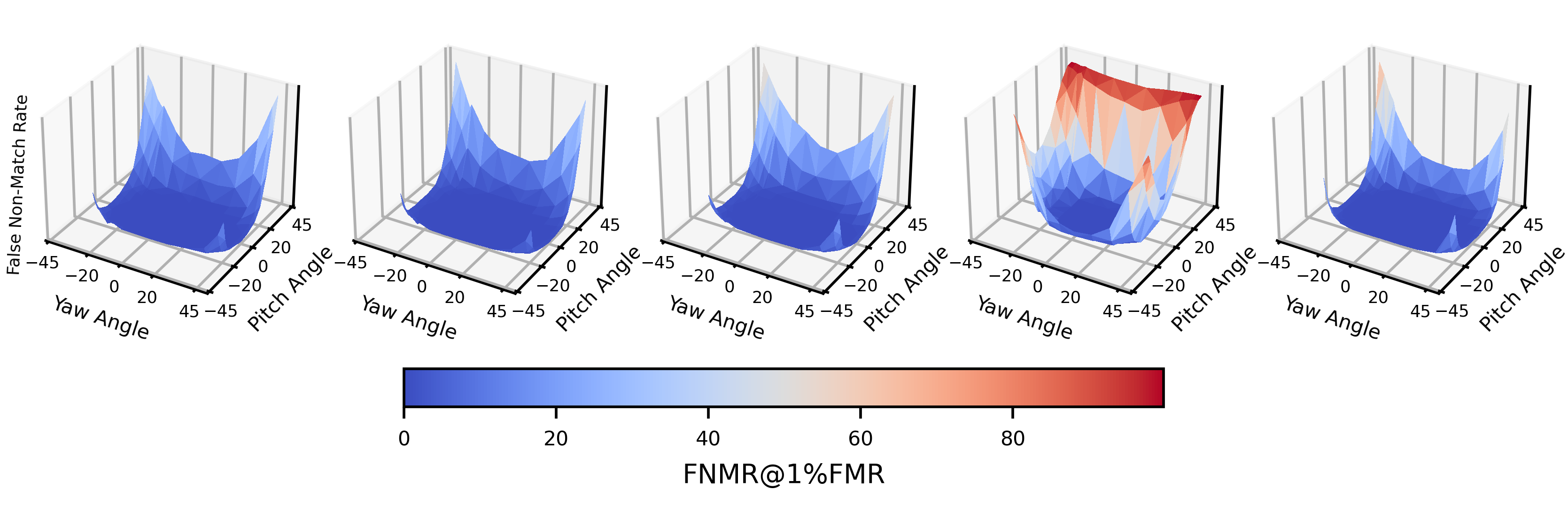}
    %\caption{StyleGAN2 \cite{karras2020analyzing}}
    \label{fig:synth-face-example}
  \end{subfigure}}
  
 \resizebox{0.7\linewidth}{!}{%
  \begin{subfigure}[b]{\linewidth}
    \includegraphics[width=\linewidth]{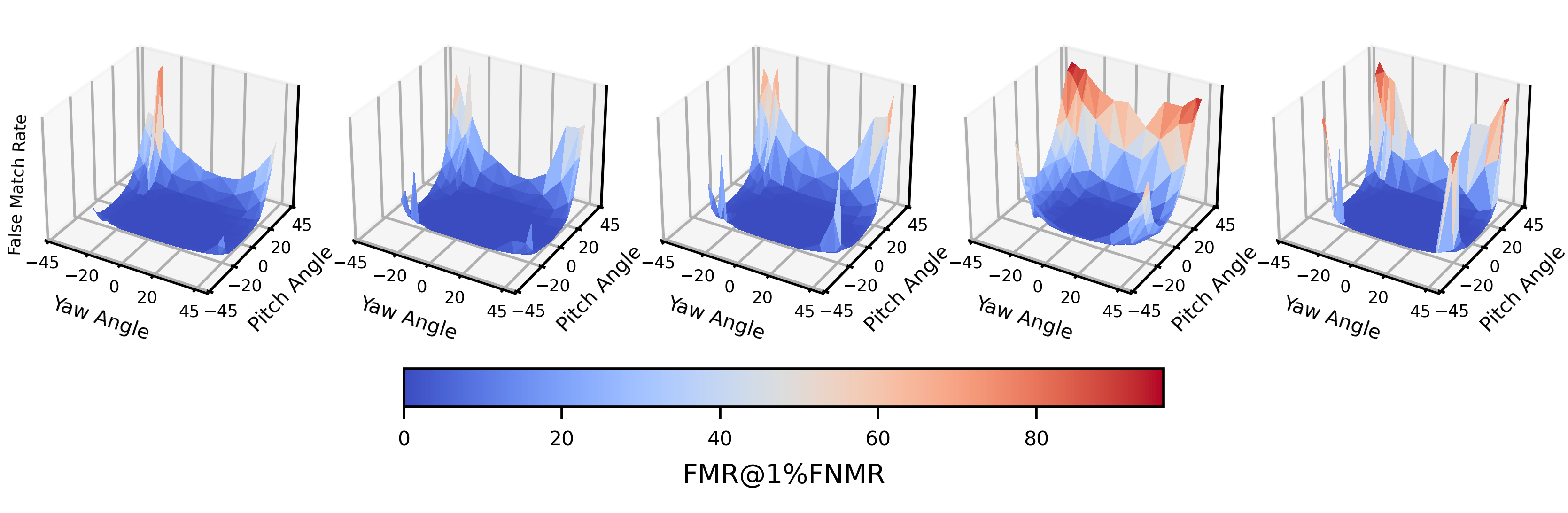}
    %\caption{StyleGAN2 \cite{karras2020analyzing}}
    \label{fig:synth-face-example}
  \end{subfigure}}

  \caption{3D Surface Grids visualising the biometric performance with respect to different yaw-pitch combinations, where each column represents one face recognition system: ArcFace (left), MagFace (center-left), CurricularFace (center), AdaFace (center-right), Cognitec (right).} \label{fig:3d-surfaces}
\end{figure*}

Figure~\ref{fig:3d-surfaces} shows the three-dimensional performance surface grid, where each point on the grid represents the performance based on 1,000 mated comparison scores obtained by comparing the frontal faces of 1,000 IDs to their mated counterparts with specific yaw-pitch combinations. This visualisation facilitates the analysis of yaw and pitch interplay and how it affects recognition performance across various metrics. \textcolor{red}{As described in Section~\ref{sec:implement-details}, we choose a truncation factor of 0.75 to randomly generate 1,000 non-mated pairs, both with frontal views. The resulting non-mated comparison score distribution is fixed throughout all our experiments for the computation of all biometric performance metrics.}

Within the interval of $[-20^{\circ}, 20^{\circ}]$, the recognition performance remains stable. However, we observe a general increase across all error metrics and FR systems for yaw and pitch angles configured to have high magnitudes simultaneously. Especially positive pitch angles beyond $30^{\circ}$ seem to cause a large performance deterioration as indicated by FNMRs and FMRs beyond $80\%$. Despite our initial expectation that some of the FR systems might excel at handling variation in pitch angles while others might perform better at handling yaw angles, our findings indicate that all FR systems consistently perform better at handling yaw angle variations. This observation lets us consider that unless all (general-purpose) FR systems are inherently biased towards variation in yaw angles, they have difficulty recognizing face images with equally high pitch angles due to limited face regions being visible.

% Synth vs Real samples

% Synth vs Real samples
\begin{figure*}
\label{fig:geometry-consistency}

\centering
\resizebox{0.7\linewidth}{!}{%
  \begin{subfigure}[b]{\linewidth}
    \includegraphics[width=\linewidth]{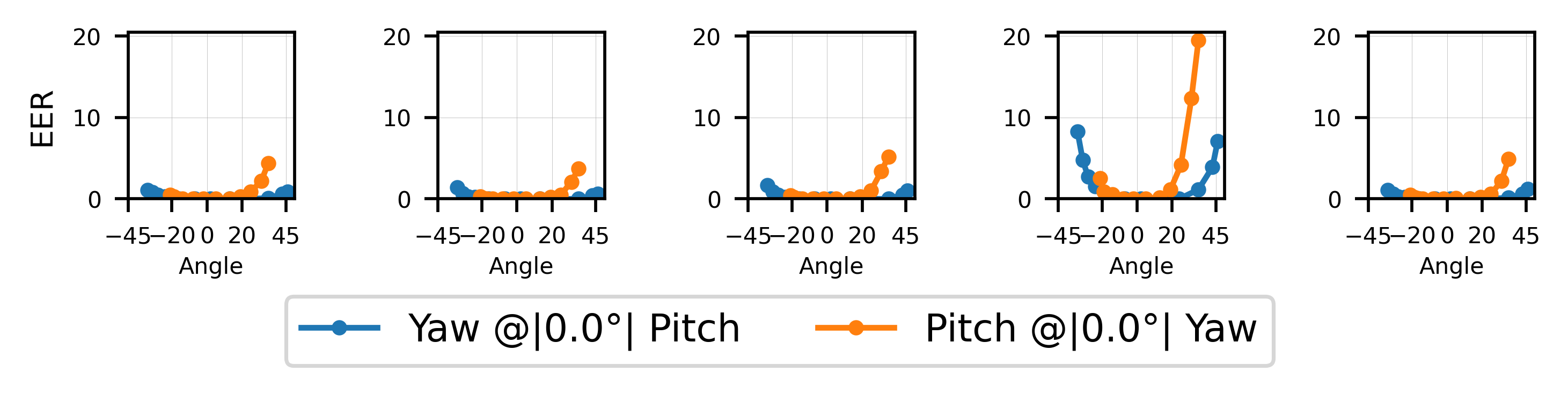}
    %\caption{StyleGAN2 \cite{karras2020analyzing}}
    \label{fig:synth-face-example}
  \end{subfigure}}
  \caption{Intersection curves of the 3D performance surface grid, showing the EERs when fixing pitch or yaw at $0^{\circ}$ for four FR systems: ArcFace (left), MagFace (center-left), CurricularFace (center), AdaFace (center-right), Cognitec (right). Pitch angles above $30^{\circ}$ cause the errors to increase significantly.}   \label{fig:2d-lines}
\end{figure*}

To further analyse the imbalances, Figure \ref{fig:2d-lines} depicts the intersection curves as a result of fixing either pitch or yaw angles at $0^{\circ}$ while iterating through the counterparts. As the view in Figure~\ref{fig:3d-surfaces} suggests, positive pitch angles beyond $30^{\circ}$ have the most significant effect on the recognition performance, as can be seen by the striking out orange curve. Although negative pitch angles could only be generated down to $-21^{\circ}$, we expect this type of variation to have a similarly high performance impact as indicated by the slight curve deflection starting in the left tail region. Positive and negative yaw angles have proven comparable, which follows our expectation as biases toward one side can be alleviated with a simple horizontal flip. 

The same imbalances can be seen in Table~\ref{tab:lasso-model-coefs}, showing the coefficients as a result of training SYP-Lasso (see Equation \ref{eq:lasso-opt}) over the four base covariates $\Phi_{p}$ and $\Phi_{y}$. The table demonstrates the overweighted negative effect of pitch angle variations, as indicated by the high magnitudes of their coefficients. In contrast, positive and negative yaw angles have a much less significant impact on the similarity scores. Also, the high adjusted $R^2$ values of up to $83\%$ express a strong linear relationship between similarity score and head pose rotation in the Syn-YawPitch dataset. 

Following the observation in this section, FR system operators should take into account the performance sensitivity of variation in pitch angles beyond $\pm30^{\circ}$. Further, we want to highlight the robustness of current open source FR systems to handle yaw-pitch intersections within $\pm20^{\circ}$, well demonstrating the capability of deep neural networks~\cite{he2016deep}\cite{behrmann2019invertible} to generalise across various camera views.

\section{Pose Quality Estimation}
\label{sec:pose-quality-estimation}

As Table~\ref{tab:lasso-model-coefs} demonstrates a strong linear relationship between head poses and similarity scores, this section analyses the effectiveness of our SYP-Lasso model for estimating the FR utility based on variations in yaw and pitch angles. \textcolor{red}{Specifically, we fit an individual SYP-Lasso regression model for each of the evaluated face recognition systems to learn the FR-specific relationships between similarity scores and yaw-pitch angle combinations. We benchmark our approach against 1) the FR-agnostic ISO/IEC-related pose quality estimator introduced in Section~\ref{sec:baseline-predictors} and 2) four well-established unified quality assessment algorithms: FaceQnet v1~\cite{hernandez2020biometric}, SER-FIQ~\cite{terhorst2020ser}, MagFace~\cite{meng2021magface}, and CR-FIQA~\cite{boutros2023cr}.}

\begin{figure}
\centering
\includegraphics[width=0.8\linewidth]{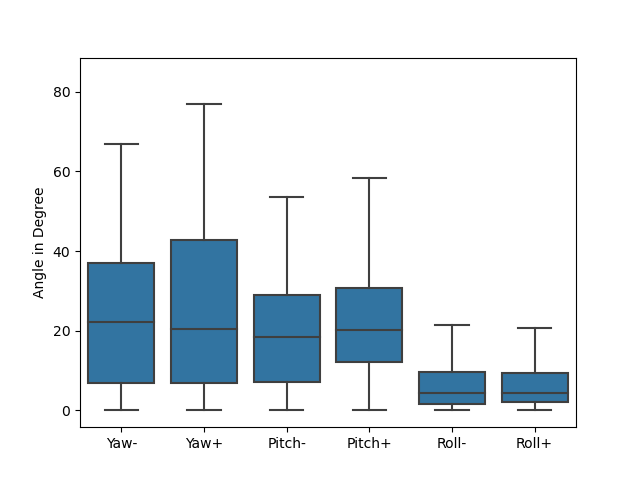}
\caption{Boxplots visualising the distribution of yaw, pitch, and roll angles in the BIWI database with negative angles reported as absolute values.    \label{fig:biwi-pose-angle-boxplots}}
\end{figure}

We base our experiments on the \textit{BIWI Kinect Head Pose} (BIWI) database~\cite{fanelli2011real} that contains $15k$ images of $24$ individuals with a yaw range of $\pm75^{\circ}$ and pitch range of $\pm60^{\circ}$. All samples were collected using 3D sensors with depth information and annotated with exact pose labels. Hence, the propagation of ground truth errors caused by inaccurate labels can be excluded. 

Figure~\ref{fig:biwi-pose-angle-boxplots} shows the distributions of yaw, pitch, and roll angles in the BIWI dataset. Since the majority of roll angles were captured in the interval of $-20^{\circ}$ to $20^{\circ}$ and EG3D being limited to editing yaw and pitch angles, we discard the analysis of roll angles in this work. This decision is motivated by the observation of previous research~\cite{lu2019experimental} attributing roll angles to have a minor impact on the FR performance within the range of $\pm30^{\circ}$. 

\begin{figure}[h]
\centering
\setlength{\tabcolsep}{1pt}
\begin{tabular}{ccccc}

\includegraphics[width=.12\linewidth,valign=m]{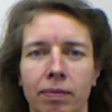} & \includegraphics[width=.12\linewidth,valign=m]{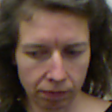} & \includegraphics[width=.12\linewidth,valign=m]{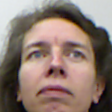} & \includegraphics[width=.12\linewidth,valign=m]{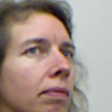} & \includegraphics[width=.12\linewidth,valign=m]{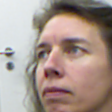} \\

\includegraphics[width=.12\linewidth,valign=m]{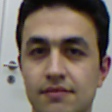} & \includegraphics[width=.12\linewidth,valign=m]{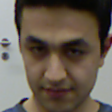} & \includegraphics[width=.12\linewidth,valign=m]{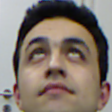} & \includegraphics[width=.12\linewidth,valign=m]{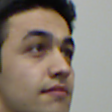} & \includegraphics[width=.12\linewidth,valign=m]{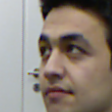} \\

\includegraphics[width=.12\linewidth,valign=m]{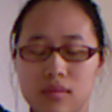} & \includegraphics[width=.12\linewidth,valign=m]{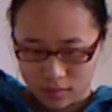} & \includegraphics[width=.12\linewidth,valign=m]{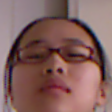} & \includegraphics[width=.12\linewidth,valign=m]{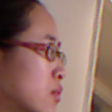} & \includegraphics[width=.12\linewidth,valign=m]{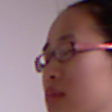} \\

\end{tabular}
\caption{Examples of selected images used to benchmark our proposed SYP-Lasso PQE against our baseline PQEs. \label{fig:biwi-imgs}}
\end{figure}

Again, we emphasise the importance of using real data to evaluate the PQE performance to exclude the \textit{synthetic-vs-real} domain gap to prevent SYP-Lasso from learning meaningful patterns. Despite the need to include real data, collecting real mated samples behaving in the same controlled manner as synthetically generated samples is not feasible without causing disproportionately high costs. The entanglement of facial attributes can be seen in Figure~\ref{fig:biwi-imgs}, which shows selected samples from the evaluation dataset. Although all samples were captured by making sure that head pose rotation constitutes the main variation factor, uncontrollable micro-expressions have an additional effect on the comparison scores.

\subsection{SYP-Lasso Benchmarking}
\label{sec:SYP-Lasso-vs-baseline}

% Synth vs Real samples
\begin{figure*}
\label{fig:geometry-consistency}

\centering
\resizebox{0.25\linewidth}{!}{%
  \begin{subfigure}[b]{\linewidth}
    \includegraphics[width=\linewidth]{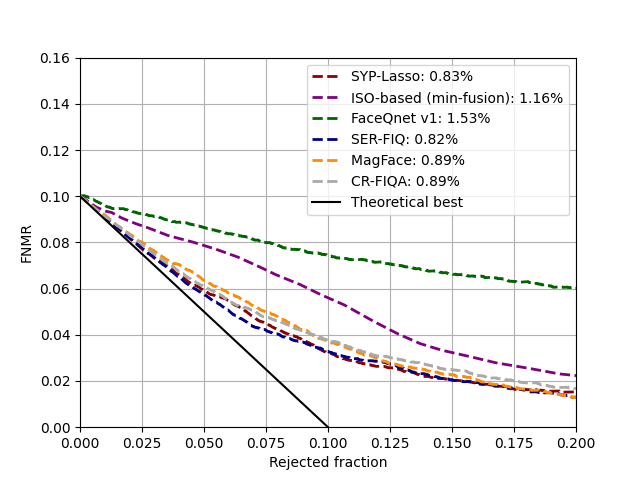}
    %\caption{StyleGAN2 \cite{karras2020analyzing}}
    \subcaption{ArcFace}
  \end{subfigure}}
  \resizebox{0.25\linewidth}{!}{%
  \begin{subfigure}[b]{\linewidth}
    \includegraphics[width=\linewidth]{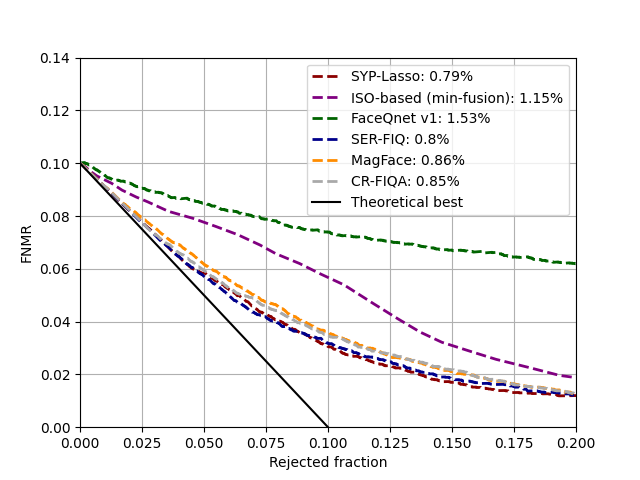}
    %\caption{StyleGAN2 \cite{karras2020analyzing}}
    \caption{MagFace}
  \end{subfigure}}
    \resizebox{0.25\linewidth}{!}{%
  \begin{subfigure}[b]{\linewidth}
    \includegraphics[width=\linewidth]{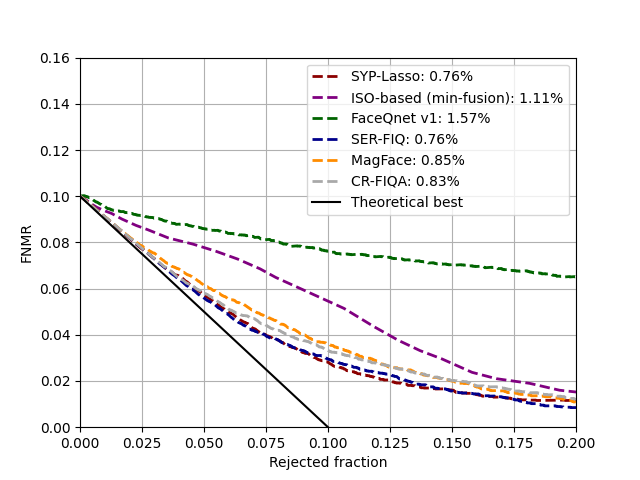}
    %\caption{StyleGAN2 \cite{karras2020analyzing}}
    \caption{CurricularFace}
    \end{subfigure}}

    \resizebox{0.25\linewidth}{!}{%
    \begin{subfigure}[b]{\linewidth}
    \includegraphics[width=\linewidth]{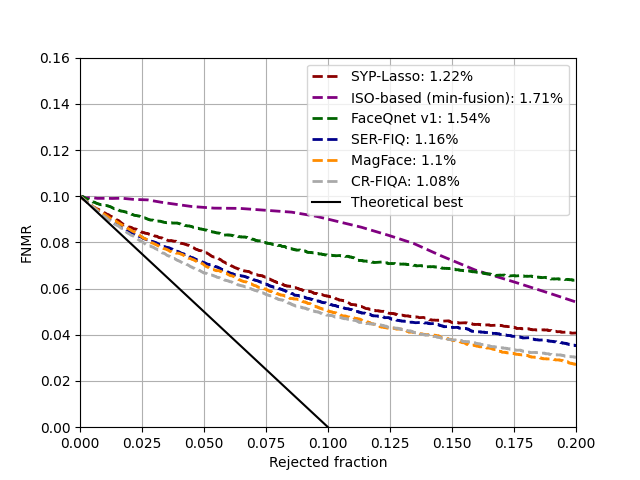}
    %\caption{StyleGAN2 \cite{karras2020analyzing}}
    \caption{AdaFace}
    \end{subfigure}}
    \resizebox{0.25\linewidth}{!}{% 
  \begin{subfigure}[b]{\linewidth}
    \includegraphics[width=\linewidth]{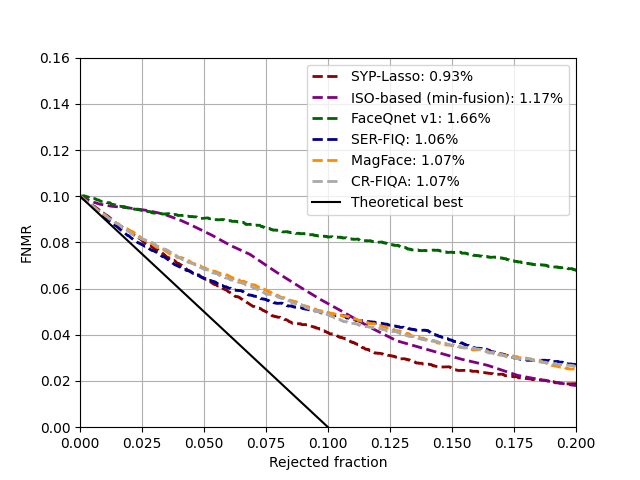}
    %\caption{StyleGAN2 \cite{karras2020analyzing}}
    \caption{Cognitec}
  \end{subfigure}}

  \caption{Error-vs-Discard Characteristic curves indicate a significant performance boost when replacing the baseline and ISO-based PQEs with SYP-Lasso. The pAUC values reported in the legend enable the quantitative comparison of the curves. We highlight the superior performance of SYP-Lasso and hence demonstrate the representativeness of our Syn-YawPitch database to data captured in the real world.}  \label{fig:edc-curves}
\end{figure*}

\textcolor{red}{We follow the specification of \textit{ISO/IEC FDIS 29794-1}~\cite{ISO-IEC-29794-1-FDIS-FaceQuality-231013} and the \textit{National Institute of Science and Technology} (NIST)~\cite{grother2021ongoing} by evaluating the performance of our SYP-Lasso quality estimator based on \textit{Error-vs-Discard Characteristic} (EDC) curves depicted in Figure~\ref{fig:edc-curves}}. Each curve visualises the FNMR after continuously discarding images with the lowest estimated pose quality. In this context, the best possible curve corresponds to the FNMRs decreasing proportionally to the percentage of discarded low-quality images. For better comparability of the curves, their \textit{partial area under curves} (pAUC) are reported up to a discard rate of $20\%$~\cite{schlett2022face}~\cite{schlett2023considerations}.

From Figure~\ref{fig:edc-curves}, it becomes evident that SYP-Lasso exhibits a significant performance advantage over the ISO/IEC-based baseline across all of the evaluated FR systems, as indicated by the lower pAUCs. Moreover, SYP-Lasso achieves competitive results with state-of-the-art unified quality assessment algorithms and even surpasses their performance for three out of five FR systems. We emphasize that the unified quality measures take into account all quality elements affecting the recognition performance, including demographic and non-demographic biases~\cite{terhorst2020face} inherited from the underlying FR system. In contrast, component quality measures focus on studying the impact of an individual quality element.

In conclusion, we demonstrate the superiority of our proposed SYP-Lasso model when compared to the existing pose quality estimation algorithm inspired by ISO/IEC CD3 29794-5~\cite{ISO-IEC-29794-5-CD3-FaceQuality-231018}. Furthermore, we achieve competitive results to established unified quality assessment algorithms in the realm of pose quality estimation while 1) utilizing only a small fraction of their model parameters and 2) yielding explainable quality measures through the application of a simple regression model.

\section{Conclusion}
\label{sec:conclusion}

In this work, we introduced Syn-YawPitch, a synthetic dataset generated with a 3D-aware GAN \cite{chan2022efficient} developed to preserve facial geometry when editing yaw and pitch angles. Further, we have analysed the influence of yaw and pitch on four open-source and one COTS FR system. Our analysis reveals that positive pitch rotations beyond $30^{\circ}$ significantly decrease the biometric performance of all FR systems. We argue that this tendency is caused by facial parts being less visible to the recognition system.

The interrelation between head pose and mated comparison scores is distilled into a lightweight Lasso regression model that has proven to outperform its ISO/IEC-based baseline, such as several state-of-the-art unified quality assessment algorithms. With SYP-Lasso being openly available to the research community, we envision future usage as foreseen by the current working draft of \textit{ISO/IEC CD3 29794-5}~\cite{ISO-IEC-29794-5-CD3-FaceQuality-231018}, going one step ahead towards more explainable face image utility measures. A future study might focus on developing a generalised PQE, as the behaviour of the evaluated FR systems stays consistent across various yaw-pitch combinations.

% or


%\appendices

%\section{}

\ifCLASSOPTIONcompsoc
  % The Biometrics Council usually uses the plural form
\section*{Acknowledgments}
\else
  \section*{Acknowledgment}
\fi
This research work has been supported by the German Federal Ministry of Education and Research and the Hessian Ministry of Higher Education, Research, Science and the Arts within their joint support of the National Research Center for Applied Cybersecurity ATHENE.

\ifCLASSOPTIONcaptionsoff
  \newpage
\fi



%\bibliographystyle{IEEEtran}
%\begin{thebibliography}{1}

%T.~Karras and P.~W. Daly, \emph{A Guide to \LaTeX}, 3rd~ed.\hskip 1em plus
%\end{thebibliography}
  
\bibliographystyle{ieeetr}
\bibliography{ref}

\end{document}